\documentclass{article}

\RequirePackage{fix-cm}

\usepackage{arxiv}

\usepackage[utf8]{inputenc} 
\usepackage[T1]{fontenc}    
\usepackage{hyperref}       
\usepackage{booktabs}       
\usepackage{graphicx}

\usepackage{tabularx}
\usepackage{multirow}
\usepackage{array}
\usepackage{subcaption}
\usepackage[shortlabels]{enumitem}

\usepackage[table,xcdraw]{xcolor}

\usepackage{caption}
\newcolumntype{Y}{>{\centering\arraybackslash}X}

\newcommand{\specialcell}[2][l]{%
  \def\arraystretch{1}
  \begin{tabular}[#1]{@{}l@{}}#2\end{tabular}}

\title{Fusing CNNs and statistical indicators to improve image classification}

\author{
 Javier Huertas-Tato \\
  Universidad Politecnica de Madrid\\
  Madrid, Spain\\
  \texttt{javier.huertas.tato@upm.es} \\
   \And
 Alejandro Mart\'in \\
  Universidad Politecnica de Madrid                                  \\
  Madrid, Spain\\
  \texttt{alejandro.martin@upm.es}      
 \AND

  Julian Fierrez \\
 Universidad Autonoma de Madrid\\
              Madrid, Spain\\
              \texttt{julian.fierrez@uam.es} \\
 \And 
 David Camacho \\
  Universidad Politecnica de Madrid\\
              Madrid, Spain\\
              \texttt{david.camacho@upm.es}
}

\begin{document}
\maketitle
\begin{abstract}
Convolutional Networks have dominated the field of computer vision for the last ten years, exhibiting extremely powerful feature extraction capabilities and outstanding classification performance. The main strategy to prolong this trend relies on further upscaling networks in size. However, costs increase rapidly while performance improvements may be marginal. Our main hypothesis is that adding additional sources of information can help to increase performance and that this approach may be more cost-effective than building bigger networks, which involve higher training time, larger parametrization space and high computational resources needs. In this paper, an ensemble method is proposed for accurate image classification, fusing automatically detected features through a Convolutional Neural Network and a set of manually defined statistical indicators. Through a combination of the predictions of a CNN and a secondary classifier trained on statistical features, better classification performance can be cheaply achieved. We test five different CNN architectures and multiple learning algorithms on a diverse number of datasets to validate our proposal. According to the results, the inclusion of additional indicators and an ensemble classification approach helps to increase the performance in all datasets. Both code and datasets are publicly available via GitHub at: \url{https://github.com/jahuerta92/cnn-prob-ensemble}.
\end{abstract}


\section{Introduction}\label{sec:introduction}

    Convolutional Neural Networks~\cite{lecun1995convolutional} have positioned themselves as an important referent in varied and well known tasks. Taking the animal virtual cortex as source of inspiration, these deep architectures extract patterns from data with great ease, greatly improving upon previous machine learning based approaches. These architectures have been widely used in academic and commercial contexts with great success and in multiple domains such as speech recognition or computer vision, being very powerful on a plethora of tasks~\cite{liu2017survey}. Given their power and wide range of application, there is a growing trend towards developing stronger architectures, which frequently consist in making wider, deeper or denser networks. This tendency is repeatedly shown in the most recent approaches. Despite not being a convolutional network, a concerning example on this trend is GPT-3~\cite{Brown2020May}, a deep language model so large (175 billion parameters) that it requires access to a large supercomputer to be trained.
    
    Some of the most used convolutional architectures such as VGG-16 and VGG-19~\cite{simonyan2014very} employ around 140 million parameters and 25 layers of depth which, depending on the domain, can take weeks of processing. Other models tend to prefer deeper architectures with a lower number of parameters. This is the case of Inception v3~\cite{Szegedy2015Dec}, which employs 25 million parameters and 150 layers. AmoebaNet~\cite{Real2018Feb} is another recent example of growing parameter space size, a model that requires great increases in network size to achieve a significant improvement, from 90 million to 470 million parameters in order to achieve a 1.1\% improvement in accuracy in the ImageNet problem. 
    
    Although building deeper and heavier architectures allows to constantly overcome previously achieved scores, we are in fact wasting the opportunity of leveraging additional well-known mechanisms to improve current results, in an efficient manner, and increasing explanatory power. While CNNs are powerful methods with excellent feature extraction capabilities, their capabilities for explanation are limited compared to, for example, decision trees, k-Nearest Neighbours and so on. On the other hand, these explainable techniques are far less powerful than CNNs in terms of recognition. Decision trees, SVMs, among others, rely on manually extracted features to infer knowledge, which offer rich information but usually incomplete and which lacks an spatial understanding of the image itself. These models are very limited in complex domains where shape, locality and other characteristics are key to inference. Therefore, a balance must be found between performance and explainability.
    
    The key to understanding this trade-off is that CNNs and feature methods employ different types of reasoning for inference. CNNs seek high performance automatising the feature extraction process, while methods focused on manually extracted features are explainable. While different, these methods can be complementary, building a fruitful combination that can help increasing performance while maintaining a level of explainability by using feature based methods. In this research, we present a curated list of statistical indicators that, once combined with the output of a CNN model through an ensemble approach, provide useful additional features that serve towards improving image classification performance. The main contributions of this research can be summarized as follows:
    
    \begin{itemize}
        \item A curated list of statistical indicators than allow to extract useful differentiating features from images.
        \item An analysis of five well-known CNN architectures and its performance in 10 different datasets.
        \item An ensemble approach to combine CNN architectures with classifiers trained on statistical indicators to improve performance in image classification tasks.
        \item A thorough experimentation, making use of 10 different image classification datasets to validate the ensemble.
        \item An ablation study to assess the performance of each individual statistical indicator in all the combinations of datasets and CNN architectures tested.
    \end{itemize}
    
    Finally, this manuscript is organised as follows: Section~\ref{sec:related_work} presents a summary of the state-of-the-art literature, with special emphasis being placed on statistical features for image classification; Section~\ref{sec:method} defines our proposal for combining statistical features and CNN architectures; Section~\ref{sec:tech_details} presents the different datasets used in the experiments, the image preprocessing steps applied, the CNN architectures tested, the hyperparameters used to train these architectures and the ML methods, details regarding the execution environment, and the link the repository with all the code. Section~\ref{sec:experimentation} shows the experimental results and Section~\ref{section:conclusions} outlines the main conclusions and potential lines of future work risen from this work.
    
\section{Related work}
\label{sec:related_work}
    Upscaling architectures is one of the most frequent approaches to increase the performance of neural networks. However, as parameters increase, so does the computational and time cost of the training process. This fact highlights the importance of considering alternative methods for improving the throughput of these methods. Some authors have recently started raising concerns about this trend, from environmental problems of the high performance computing required to train the larger architectures~\cite{Schwartz2019Jul} to the accessibility of Deep Learning research~\cite{strubell_energy_2020}. Although lightweight architectures are trainable by making use of standard hardware, most of the state-of-the-art models require multiple expensive GPUs resources to be trained~\cite{Steinkraus2005Aug}. Due to this, there is a growing interest in efficiently improving CNN architectures using minimal additions.
    
    
    A potential course of action to be explored is Network compression~\cite{xn--Bucilu-85b2006Aug}, which aims to reduce the size of deep neural networks while trying to maintain their effectiveness intact. Several techniques have been proposed in this direction~\cite{Cheng2017Oct,martin2020statistically}. For instance, Knowledge Distillation~\cite{Hinton2015Mar} which trains a student model with the predictions of an already existing larger (teacher) model. Another typical approach consists on pruning redundant or low-information weights~\cite{He2017,Luo2017,martin2020optimising} that are able to compress state-of-the-art CNNs to achieve at least 2x speedup with minimal losses (around 1\% accuracy). Compacted filters~\cite{Cohen2016Feb} have also been proposed aiming to optimize the size of the network by factorizing filters. For example, 3x3 convolutions can be replaced by 1x1 convolutions on SqueezeNet~\cite{Iandola2016Feb}, achieving 50x fewer parameters than AlexNet. Other approaches focus on searching for the most appropriate architecture~\cite{martin2018evodeep}.
    
    Computational efficiency is also an important research line in this scenario. Different architectures have been recently developed in this line optimising convolutional architectures. For example, EfficientNet~\cite{Tan2019May} focuses on speeding up the neural network instead of scaling the network with more parameters, offering speedups ranging from 5 to 10 compared with competitive state-of-the-art models such as AmoebaNet, NASNet~\cite{Zoph2017Jul} or GPipe~\cite{Huang2018Nov}, and maintaining similar effectiveness while reducing the number of parameters by a factor of 4 to 7. Other methods are focused on making lightweight powerful networks for mobile devices, assuming that larger networks cannot be handled as in other platforms. Two examples of this trend are ShuffleNet~\cite{Zhang2018} and MobileNet~\cite{Howard2017Apr}, focused on developing novel operators and architecture optimisation to create lighter models. 
    
    
    Information fusion~\cite{Fierrez2018-1,Fierrez2018-2} also represents an important avenue for the improvement of CNNs, which is commonly built in two distinct ways: ensembles of neural networks or appending an information fusion algorithm after the CNN representation features are extracted. An example of this would be a recent approach~\cite{Roy2020} which fuses the representation features of AlexNet~\cite{krizhevsky2017imagenet} and VGG16 to increase performance over benchmark datasets. Alternatively, it is possible to split the image into several patches and processing them separately with convolutions, to later merge the representation features~\cite{Amin-Naji2019Nov}. Another interesting application~\cite{Deng2019Nov} has been proposed in a domain where various shallow networks coexist with a PCA analysis of inputs to build a combined feature vector for biomedical information. As mentioned before, ensembles can also be built by appending an information fusion stage after the features are extracted as proposed by~\cite{Huang2006Jun}. This notion is again used by~\cite{Niu2012Apr} in their research, where another SVM is appended after the convolutional features to make decisions. In~\cite{liz2021ensembles} an ensemble of a set of deep learning models is designed to classify chest X-rays using small datasets for training, which is later combined with an explainable artificial intelligence (XAI) technique, based on combining the individual heatmaps obtained from each model in the ensemble. This approach, based on fusion and XAI methods, is used to improve both the overall performance classification of CNN and the interpretability of the CNNs, highlighting those areas of the image which are more relevant to generate the classification. Nevertheless, the use of fusion methods to improve performance has been demonstrated in plenty of domains~\cite{martin2019android}.
    
    Taking a step back, before CNNs were used, features were usually extracted by experts manually. During this process, the expert takes some input data (image pixels in computer vision) and transforms them into a manageable set of features~\cite{Nixon2012Oct,storcheus2015survey}. Once features are extracted, they can be processed through other machine learning algorithms. There are many challenges in this area~\cite{Khalid2014Aug} but, when they are properly addressed, it is possible to achieve high performance on high dimensional data. There is plenty of literature published around this issue. Originally textural features were coined by Haralick et al.~\cite{haralick_textural_1973}, who proposed a grey level co-occurrence matrix (GLCM) that contains the relative frequency of pixel pairs in an image. Further textural analysis has also been proposed, such as Local Binary Patterns (LBPs)~\cite{Ojala2002Aug}, computing comparisons on neighbouring pixels across a circle and extracting the frequency histogram of these comparisons.
    
    These techniques are paired with machine learning algorithms. Support Vector Machines (SVM)~\cite{cortes1995support} are used in several relevant works, for example in texture classification~\cite{Kim2002Nov}, on face biometrics~\cite{Sosa2018}, on-road car detection~\cite{Sun2020Oct} or handwriting recognition~\cite{Faundez2021}, among others. K-Nearest Neighbors is also used to classify features, for example on medical abnormality detection~\cite{ramteke2012automatic}, plant leaf recognition~\cite{Munisami2015Jan} or MRI image classification~\cite{Rajini2020}. Another frequent classification algorithm for image features are Random Forests (RF)~\cite{breiman_random_2001}, with relevant works at a variety of tasks such as radar image classification~\cite{Du2015Jul}, leukhemia detection~\cite{Mishra2017Mar} or brain scan classification~\cite{Gray2013Jan}.
    
    Other authors have put the focus on the information that can be extracted with different levels of granularity. Ding et al. propose the use of a two-level classification~\cite{ding2021ap}, in this case making the use of the attention mechanism and a pyramidal strategy to refine the classification with fine-grained information. In contrast, our focus is to extract general patterns using two different strategies to extract complementary information. Chang et al.~\cite{chang2020devil} also concentrate on fine-grained image classification by making use of a mutual-channel loss function that includes discriminality and diversity components. Our approach also leverages the information contained across channels, but using different statistical indicators such as average or deviation that can help to extract important patterns with maximum efficiency. Another fusion approach for image classification proposed in the literature consists in making use of the mid and high-level features extracted using a Bilinear CNN~\cite{peng2019fb}.

    In this work, a technique for the efficient improvement of image classification tasks is proposed by fusing manual feature extraction and CNNs. In the next sections, we show that predictions offered by models that rely on manually extracted features and those provided by CNNs entail complementary descriptive features, and therefore, that their combination will successfully improve both techniques. As manual features require lightweight machine learning algorithms to predict, the end performance of the ensemble will improve with a negligible increase in cost compared to the CNN computational requirements. To further improve CNNs performance, an ensemble is proposed consisting on the combination of manual feature extraction and CNNs by combining their respective end classification labels into a single label prediction~\cite{fierrez06phd}.

    \begin{figure*}  
		\centering
		\includegraphics[width=0.8  \textwidth]{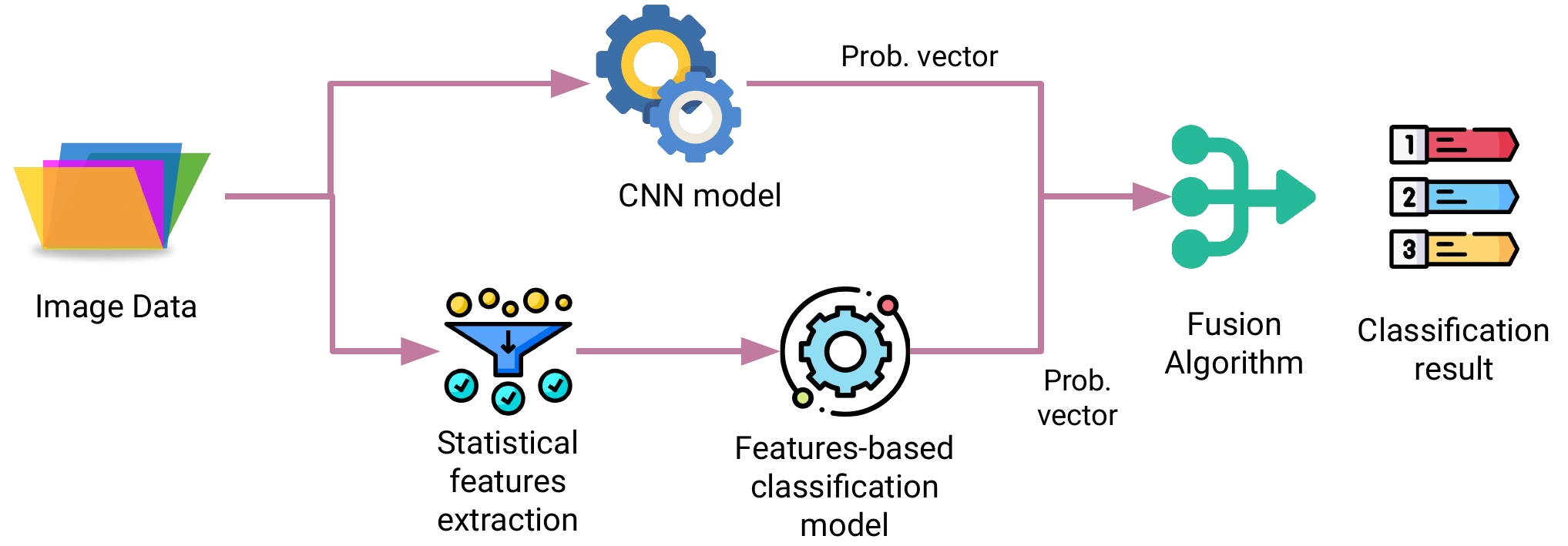}
		\caption{Visual representation of the fusion approach proposed. The images are used to train a CNN architecture. Simultaneously, different statistical indicators are extracted from every image, and a classification model is trained on these features. Finally, an ensemble is built, training a new classification algorithm that combines the outputs of the CNN and the features-based classifier.}
		\label{fig:6_algorithm_overview}
	\end{figure*}

\section{Improving CNN classification with statistical indicators
}\label{sec:method}

    With the rise of research focused on convolutional architectures, previous approaches using manually extracted features have been mostly abandoned. However, although less powerful, statistical indicators from images proved to be useful in combination with classical machine learning classifiers, reaching reasonable classification performance rates in certain domains. Besides, in contrast to the features extracted after different filters are applied in a convolutional architecture, these statistical indicators provide understandable information with minimal computational resources needed.
    
    In this research we leverage this statistical indicators to improve the performance of Convolutional Neural architectures. The core idea behind this method is to build an ensemble approach, combining a CNN model with a classical machine learning algorithm trained on the statistical information of the images, aiming to achieve improved CNN performance by adding light-weight models. In short, the method proposed combines the output probabilities of a CNN with softmax activation with the output probabilities of a classic machine learning algorithm trained on a series of statistical indicators representing different characteristics of the input images. This is meant to blend expert knowledge, providing a set of general-purpose statistical indicators and automatic feature detection provided by a powerful CNN architecture. The end goal is to significantly improve the performance of CNN architectures with minimal additional investment. 
    

    \subsection{Architecture overview}
    An overview of the solution is presented in Figure~\ref{fig:6_algorithm_overview}. Four modules are found on this system: the CNN, the statistical features extraction process, the statistical features-based classifier and the fusion algorithm, which consists of a classical machine learning algorithm used to achieve the best combination between the output probabilities delivered by the CNN and the statistical features-based classifier.
	
	In detail, to train this architecture the following steps are executed in accordance to Figure~\ref{fig:6_algorithm_overview}:
	\begin{enumerate}
	    \item From the images, a pool of general-purpose features are extracted, transforming each image into a vector of representative features.
	    \item The image features extracted in step 1 are used to train the features-based model. Using a robust method is encouraged as some of the general-purpose features could be irrelevant for some datasets, however, most methods are able to improve classification performance.
	    \item In parallel to steps 1 and 2 and using the same images as before, a CNN is trained. Hyper-parameters and the optimal topology has to be adjusted as usual, aiming for high performance on a validation dataset.
	    \item The output probabilities of step 2 and step 3 are concatenated for each processed image. All images from the training set are transformed into concatenated vectors of probabilities. If a machine learning algorithm does not support probabilities, the one-hot representation of the class is enough to represent the prediction.
	    \item The concatenated probabilities are used to train a fusion learning algorithm. It is worth noting that if optimal results are to be achieved several models have to be tested on steps 3 and 5, but improving results upon the original CNN does not require such experimentation. Additionally, we also test a simple average between both vectors.
	    \item A vector is obtained with the final classification.
	\end{enumerate}
	
	The following sections describe the methods tested to fulfill the CNN, Feature and Fusion algorithm modules of the architecture. Our motivation is to show that, in most of the cases, a CNN complemented with any of the proposed features and fusion algorithm achieves better performance on classification tasks than the CNN alone.
	
	\subsection{Image classification with CNNs}
        As shown in Figure~\ref{fig:6_algorithm_overview}, one of the classification methods that composes the architecture proposed is a CNN, which takes the raw image and automatically extracts features through different convolutional layers and infers knowledge in the last layers of the network. The fusion method proposed is independent of the CNN architecture, so it can be used with any convolutional network. Once the CNN is trained on a training set, the output probabilities of the model are combined with the output of a classifier trained on the statistical information through a machine learning classifier (or the average between probability vectors). As we will describe in Section~\ref{sec:tech_details}, we tested the approach in five different state-of-the-art CNN architectures, including DenseNet-201, Inception-ResNet-V2, Inception-V3, VGG19 and Xception V1.

    \subsection{Extraction of statistical indicators}
    \label{sec:statistical_indicators}
        Manual extraction of features from images to perform computer vision tasks has remained a common approach in the state-of-the-art literature, although not as prolific as CNNs automatic feature extraction. On applied domains of application simple statistical features are still relevant \cite{heinle_automatic_2010,Pu_Sun_Ma_Cheng_2015,Tuominen_Pekkarinen_2005,Venkataraman_Mangayarkarasi_2016}. In this research, we propose a general-purpose statistical-based feature set for domain-independent computer vision tasks. Due to the specificity level of each of these features, their importance can vary across domains, which means that some of them may entail irrelevant knowledge in certain domains. Thus, a robust learning algorithm is used to process these statistical features, which helps to take advantage from the information offered by these features while removing superfluous information that could hinder the learning capabilities of the approach. A set of commonly used machine learning algorithms was tested, including Random Forest (RF)~\cite{breiman_random_2001}, Support Vector Machines (SVM)~\cite{cortes1995support}, Linear Discriminant Analysis (LDA), among others.
        
        Our features can be divided on two categories found in the literature about statistical features for computer vision tasks: \textit{spectral features} and \textit{textural features}~\cite{haralick_textural_1973,jensen1979spectral,pena2014object,zhang2017image}. Spectral features represent colour as an statistic, for example via average, deviation or differentials. On the other hand, textural features~\cite{haralick_textural_1973} represent an image in terms of edges and other abstract traits. This is achieved via a Grey Level Covariance Matrix (GLCM), a technique that detects the frequency of contiguous appearances of pixel values. Image textures are computed from the GLCM.
        
        Both types of features are extracted from the raw image, including colour-based information using statistical summarization of each channel (average, deviation and so on) and from Grey Level Covariance Matrix (GLCM) for each colour. The GLCM detects the frequency of contiguous appearances of pixel values, allowing the detection of textures within the image which can reveal important patterns. 
        
        More specifically, the set of features extracted includes the average, standard deviation, skewness, average colour difference, histogram (with 5 bins), average colour ratio, textural average, variance, homogeneity, contrast, dissimilarity, entropy, second movement, and correlation. Each of these features is calculated for each colour channel, summing up to 60 features. These are all described in Table~\ref{tab:features}, where the $n$-th image on a particular channel $c$ is named $I_n^c$ with pixels $I^c(i,j)$ on position $i$ and $j$ with size NxM. The corresponding probability found in the GLCM for two contiguous grey values $a$ and $b$ is $p^c(a,b)$ for Z grey levels measured. We define $p_x^c(a)$ as the $a$-th value of the sum of rows of the $p^c$ matrix.
        
        Once extracted, five different well-known classic machine learning classifiers are trained. The aim is to infer knowledge from these statistical indicators that can help to improve the performance of a CNN model. The ML algorithms, trained with the Scikit-learn~\cite{scikit-learn}, are the following:

        \begin{itemize}

            \item \textbf{K-Neighbours Classifier:} This method, in contrast to the previous ones, stores certain training instances which are later used to predict the output of a new instance. For that purpose, the label is calculated according to the neighbours of the new instance.

            \item \textbf{Linear Discriminant Analysis:} LDA is also a linear classification method which generates decision boundaries using class conditional densities and Bayes' rule.
    
            \item \textbf{Logistic Regression:} A linear classification model where the outputs are calculated according to a logistic curve and using the Broyden–Fletcher–Goldfarb–Shanno algorithm~\cite{fletcher2013practical} as optimiser.

            \item \textbf{Random Forest:} A combination of decision tree classifiers trained on sub samples of the data that averages the output of each tree to improve the accuracy.

            \item \textbf{Support Vector Machine:} This algorithm finds the hyperplane which better separates each class region to the other and maximising the separation with every data point. We have trained SVMs with two different kernels: sigmoidal and with a radial-basis function.
            
        \end{itemize}
        
        \begin{table*}
    \centering
    \scriptsize
    
    \def\arraystretch{3}
    \resizebox{0.9\linewidth}{!} {%
    			\begin{tabular}{lr}
    			\toprule
    			Average & $\mu^c=\frac{1}{NM}\sum_{i=1}^{N}\sum_{j=1}^{M}I^c(i,j)$\\
    			Standard Deviation & $\sigma^c=\sqrt{\frac{1}{NM}\sum_{i=1}^{N}\sum_{j=1}^{M}(I^c(i,j)-\mu^c)^2}$\\
    			Skewness & $\gamma^c=\frac{\frac{1}{NM}\sum_{i=1}^{N}\sum_{j=1}^{M}(I^c(i,j)-\mu^c)^3}{(\sigma^c)^3}$\\
    			Difference & $d^{c_1,c_2}=\mu^{c_1} - \mu^{c_2}$\\
    			Histogram & $h^c(B)= \sum_{i,j} 1 \textrm{ for } i,j \textrm{ subject to } b_{min} < I^c(i,j) < b_{max},$\\ 
    			          & \specialcell{where $B=\{b_{min},b_{max}\}$ is a given histogram bin, and\\ $b_{min}$ and $b_{max}$ are the minimum and maximum pixel values for a bin.\\(Only values within that bin are counted in the sum.)}\\
    			Ratio & $r^{c_1,c_2}=\frac{\mu^{c_1}}{\mu^{c_2}}$\\
    			Textural Average & $f^c_1=\sum^{2Z}_{k=2}k \{\sum^{Z}_{a=1}\sum^{Z}_{b=1}p^c(a,b)\}; a+b=k$\\ 
    			Variance & $f^c_2=\sum^{Z}_{a=1}\sum^{Z}_{b=1}(a-p_\mu^c)p^c(a,b)$\\
    			         & \specialcell{where $p_\mu^c$ is the average of $p^c$}\\
    			Homogeneity & $f^c_3=\sum_{a=1}^{Z}\sum_{b=1}^{Z}\frac{1}{1+(a-b)^2}p^c(a,b)$\\
    			Contrast & $f^c_4=\sum^{Z}_{a=1}\sum^{Z}_{b=1}p^c(a,b)(a-b)^2$\\
    			Dissimilarity &  $f^c_5=\sum^{Z}_{a=1}\sum^{Z}_{b=1}p^c(a,b)|a-b|$\\
    			Entropy &$f^c_6=-\sum^{Z}_{a=1}\sum^{Z}_{b=1}p^c(a,b)\log(p^c(a,b))$ \\
    			Second Moment &  $f^c_7=\sum_{a=1}^{Z}\sum_{b=1}^{Z}p^c(a,b)^2$\\
    			Correlation & $f^c_8=\frac{\sum_{a=1}^{Z}\sum_{b=1}^{Z}(ij)p^c(a,b)-p_{x,\mu}^cp_{y,\mu}^c}
    			              {p_{x,\sigma}^cp_{y,\sigma}^c}$\\
    			            &\specialcell{where $p_{x,\mu}^c,p_{y,\mu}^c,p_{x,\sigma}^c,p_{y,\sigma}^c$ are the averages \\and standard deviations of $p_x^c, p_y^c$.}\\
    		    \bottomrule
    \end{tabular}
    } 
    
    \caption{List of features and formulas used for the proposed hand-crafted feature extraction.}
    \label{tab:features}
\end{table*}

    \subsection{Fusion of CNN and statistical-based classification output probabilities}
    
        The result of training a CNN architecture in a given domain is a model where the last fully connected layer, for a certain input instance, delivers a vector defining a set of probabilities for each class. Normally, the maximum argument in this vector indicates the final label. However, classes could also reach similar probabilities, causing a decision without enough certainty. We argue that the inclusion of the statistical indicators previously described can help to achieve more accurate decisions in these cases, leading to an improvement of the final classification performance.
        
        The final classification in our proposal is the result of the combination of two probability vectors: (a) the classification probabilities according to the CNN model, and (b) the output probabilities of a ML algorithm trained with the statistical indicators extracted. If the ML algorithm finally selected only provides a label instead of class probabilities, a one hot-encoding representation is used. After the modules for image and feature classification have been selected, a final module is required to combine the outputs of both sources. For that purpose, the probabilities of an image are calculated for both classifiers, the output vectors are concatenated and, finally, act as training examples for a fusion algorithm. For this final method, it is used the same set classifiers used for the statistical features-based classifier (described in Section~\ref{sec:statistical_indicators}). The fusion algorithm outputs the end label from the original image.

\section{Experimental setup}
\label{sec:tech_details}

This section describes in detail the data, CNN architectures tested, image preprocessing procedure and parameters used during the experimentation. 

    \subsection{Datasets}
    \label{sec:data}
    
    \begin{figure}[!htpb]
    \centering

     \begin{subfigure}[b]{0.35\textwidth}
         \centering
        \includegraphics[width=0.28\textwidth]{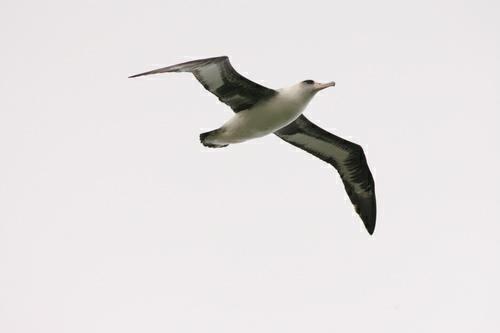}
        \includegraphics[width=0.28\textwidth]{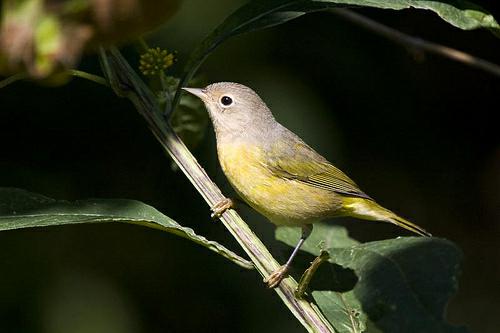}
        \includegraphics[width=0.28\textwidth]{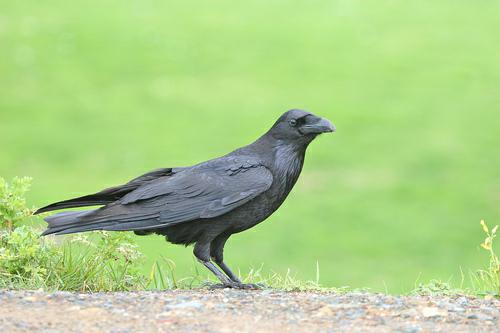}
        \caption{Caltech birds2011}
        \label{fig:caltech_birds2011}
         
     \end{subfigure}

     \begin{subfigure}[b]{0.35\textwidth}
         \centering
        \includegraphics[width=0.28\textwidth]{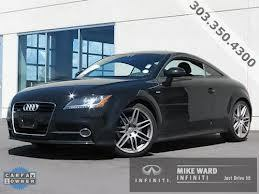}
        \includegraphics[width=0.28\textwidth]{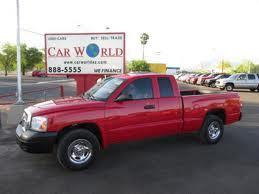}
        \includegraphics[width=0.28\textwidth]{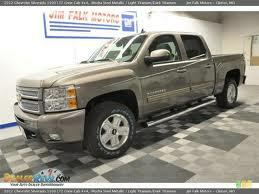}
        \caption{Cars196}
        \label{fig:cars196}
     \end{subfigure}
     
    \begin{subfigure}[b]{0.35\textwidth}
         \centering
        \includegraphics[width=0.28\textwidth]{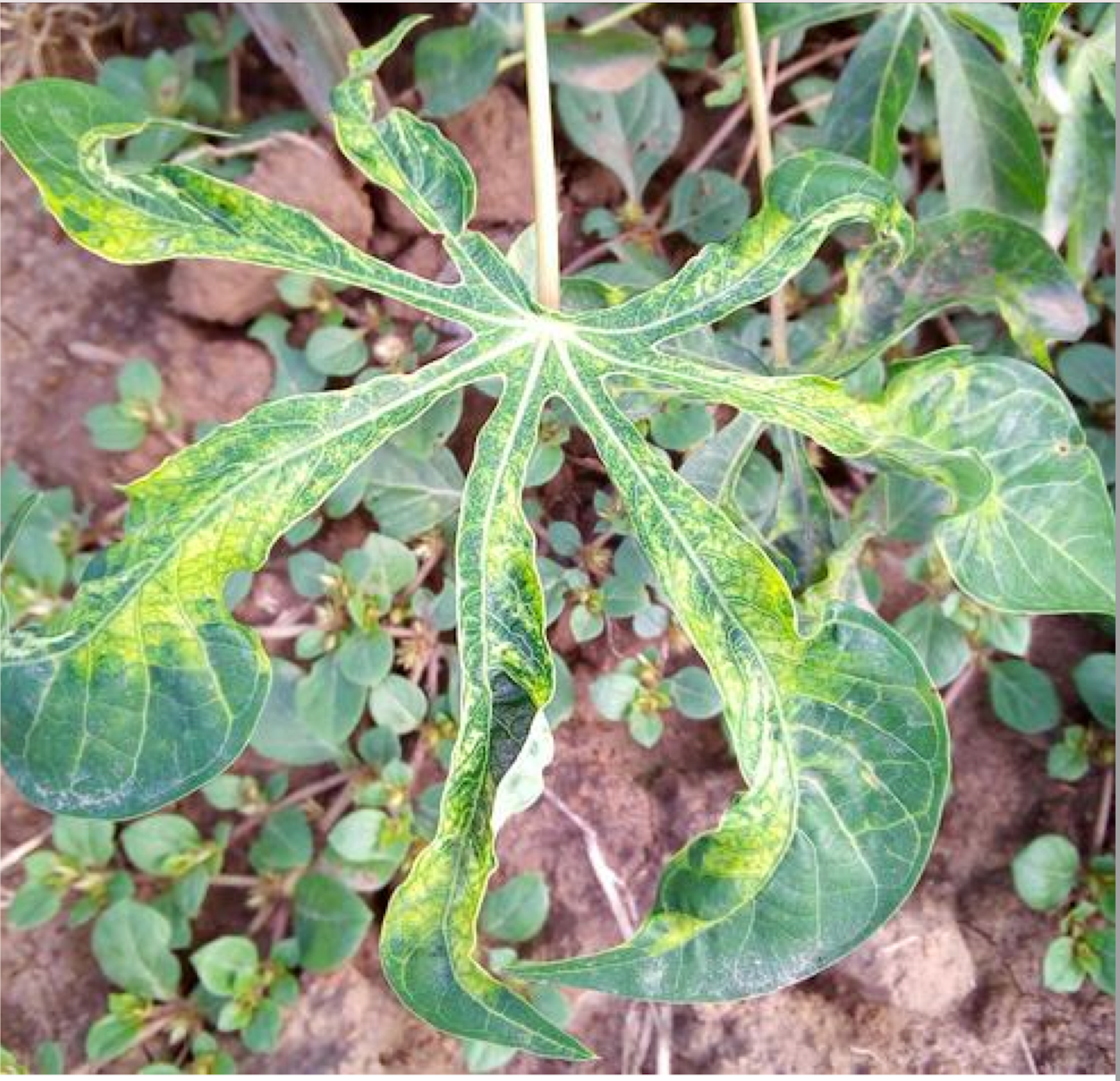}
        \includegraphics[width=0.28\textwidth]{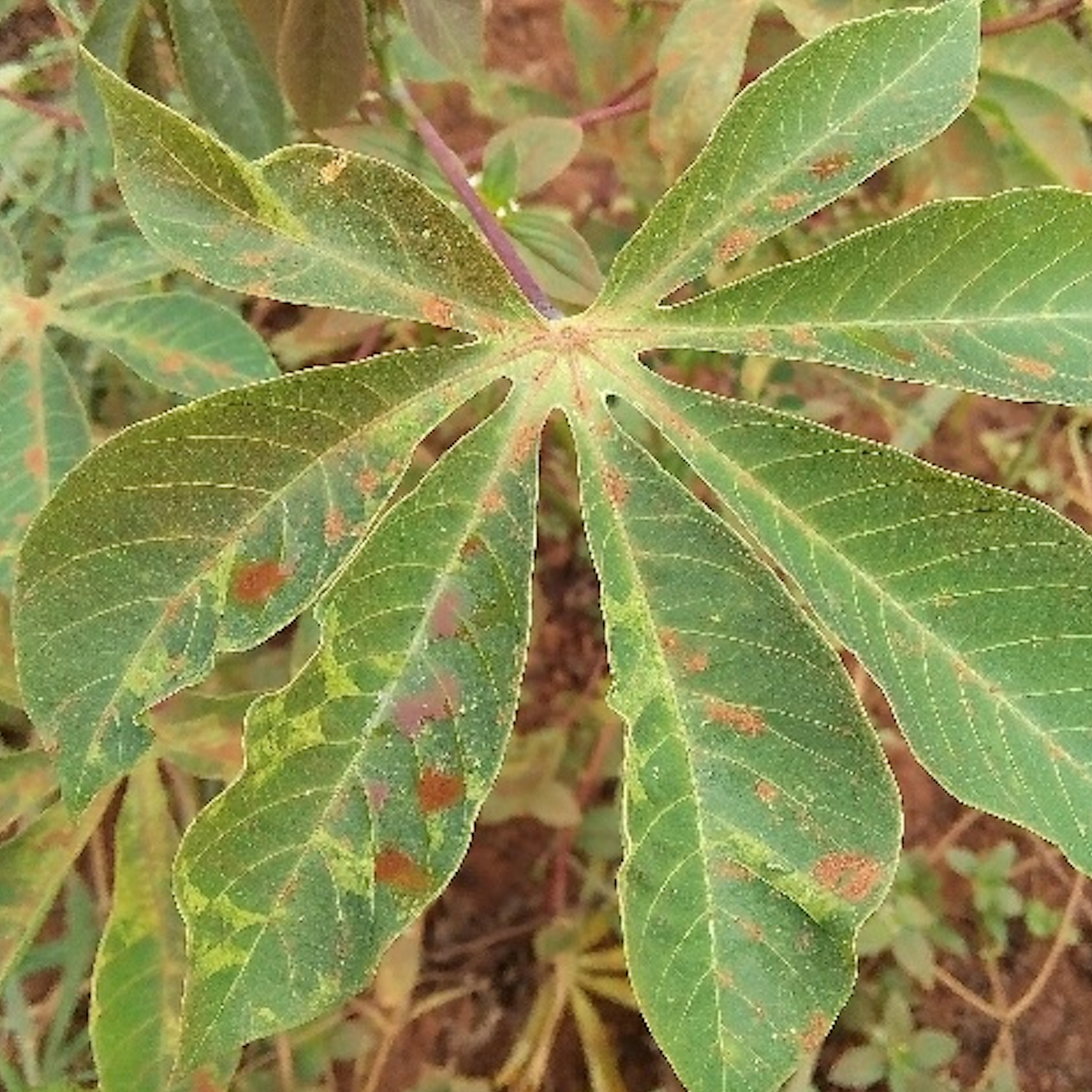}
        \includegraphics[width=0.28\textwidth]{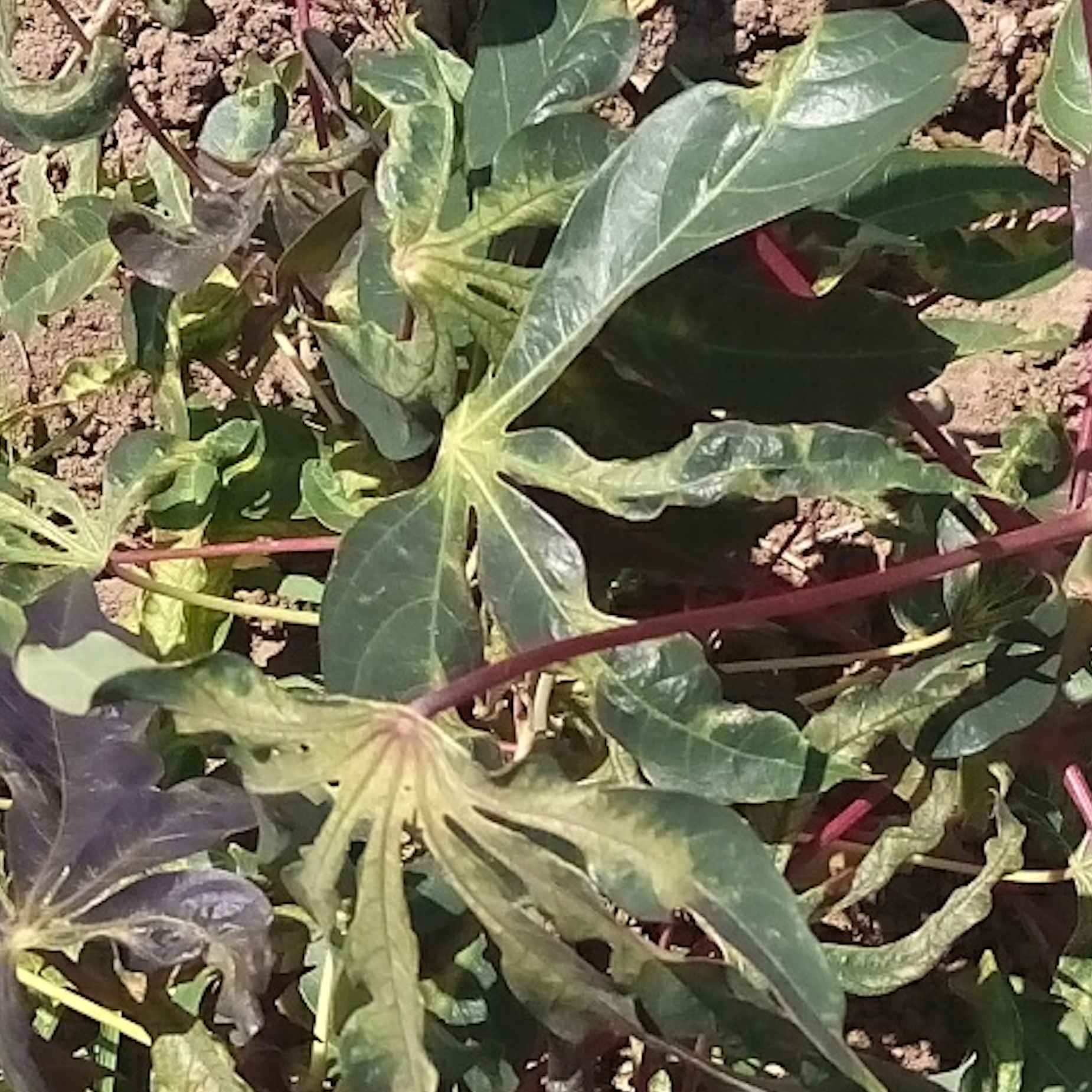}
        \caption{Cassava}
        \label{fig:cassava}
     \end{subfigure}
     
         \begin{subfigure}[b]{0.35\textwidth}
         \centering
        \includegraphics[width=0.28\textwidth]{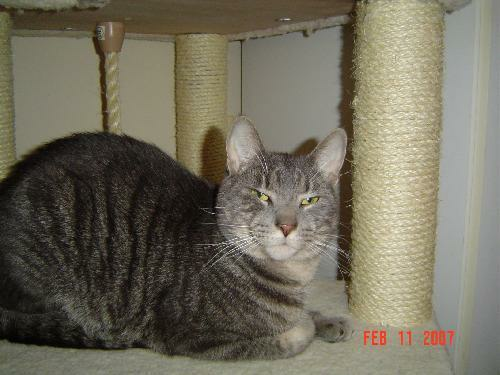}
        \includegraphics[width=0.28\textwidth]{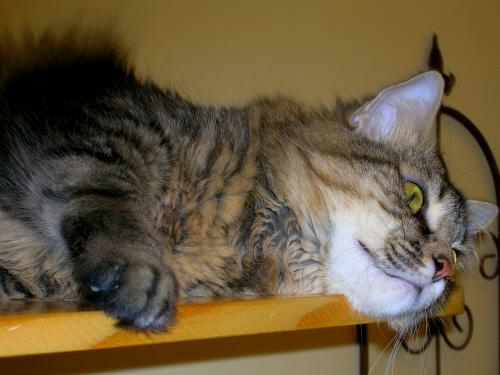}
        \includegraphics[width=0.28\textwidth]{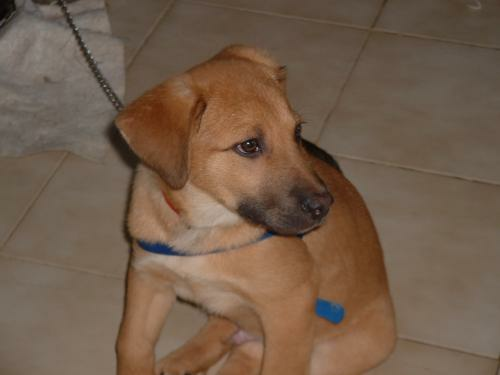}
        \caption{Cats vs dogs}
        \label{fig:cats_vs_dogs}
     \end{subfigure}

         \begin{subfigure}[b]{0.35\textwidth}
         \centering
        \includegraphics[width=0.28\textwidth]{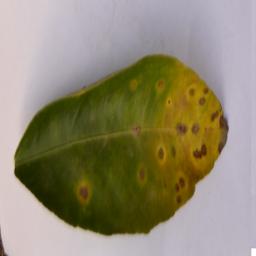}
        \includegraphics[width=0.28\textwidth]{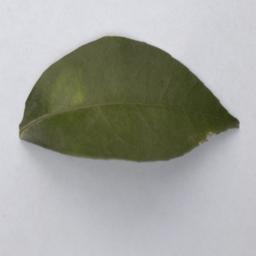}
        \includegraphics[width=0.28\textwidth]{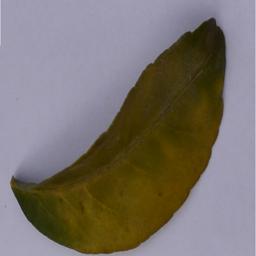}
        \caption{Citrus leaves}
        \label{fig:citrus_leaves}
     \end{subfigure}
     
     \begin{subfigure}[b]{0.35\textwidth}
         \centering
        \includegraphics[width=0.28\textwidth]{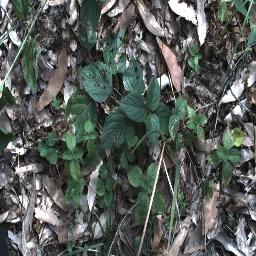}
        \includegraphics[width=0.28\textwidth]{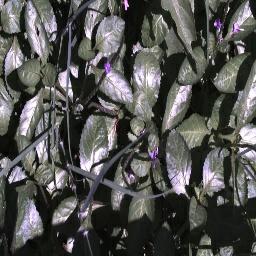}
        \includegraphics[width=0.28\textwidth]{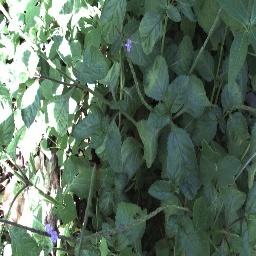}
        \caption{Deep weeds}
        \label{fig:deep_weeds}
     \end{subfigure}
     
        \begin{subfigure}[b]{0.35\textwidth}
         \centering
        \includegraphics[width=0.28\textwidth]{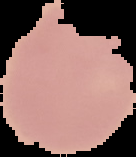}
        \includegraphics[width=0.28\textwidth]{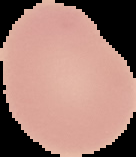}
        \includegraphics[width=0.28\textwidth]{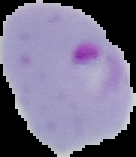}
        \caption{Malaria}
        \label{fig:malaria}
     \end{subfigure}

        \begin{subfigure}[b]{0.35\textwidth}
         \centering
        \includegraphics[width=0.28\textwidth]{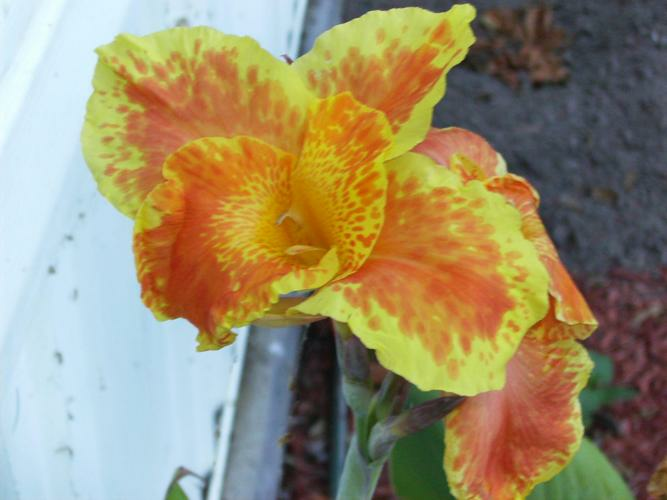}
        \includegraphics[width=0.28\textwidth]{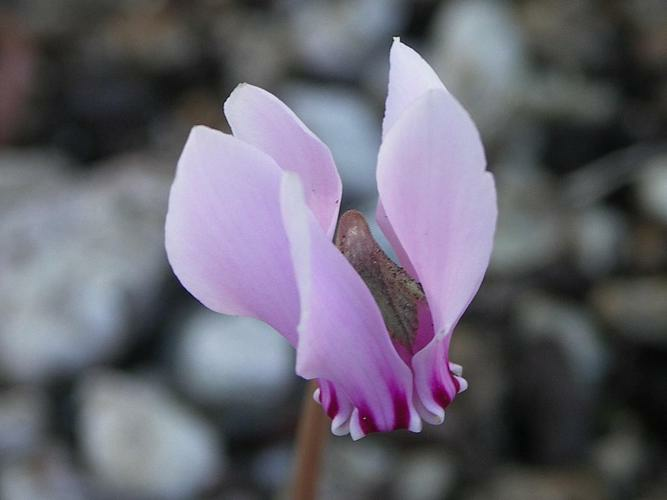}
        \includegraphics[width=0.28\textwidth]{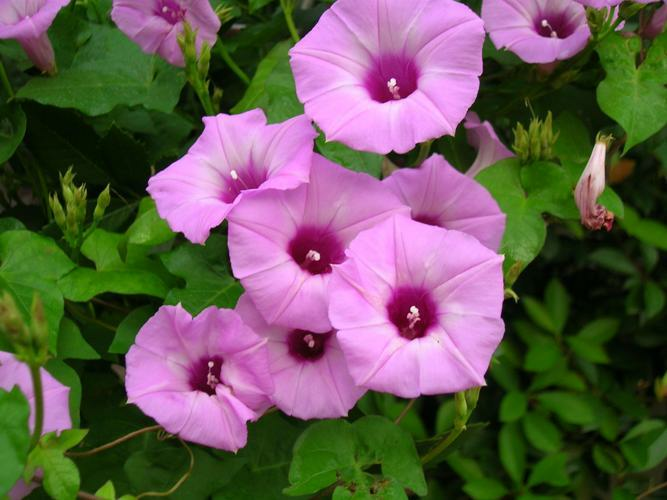}
        \caption{Oxford flowers102}
        \label{fig:oxford_flowers102}
     \end{subfigure}

        \begin{subfigure}[b]{0.35\textwidth}
         \centering
        \includegraphics[width=0.28\textwidth]{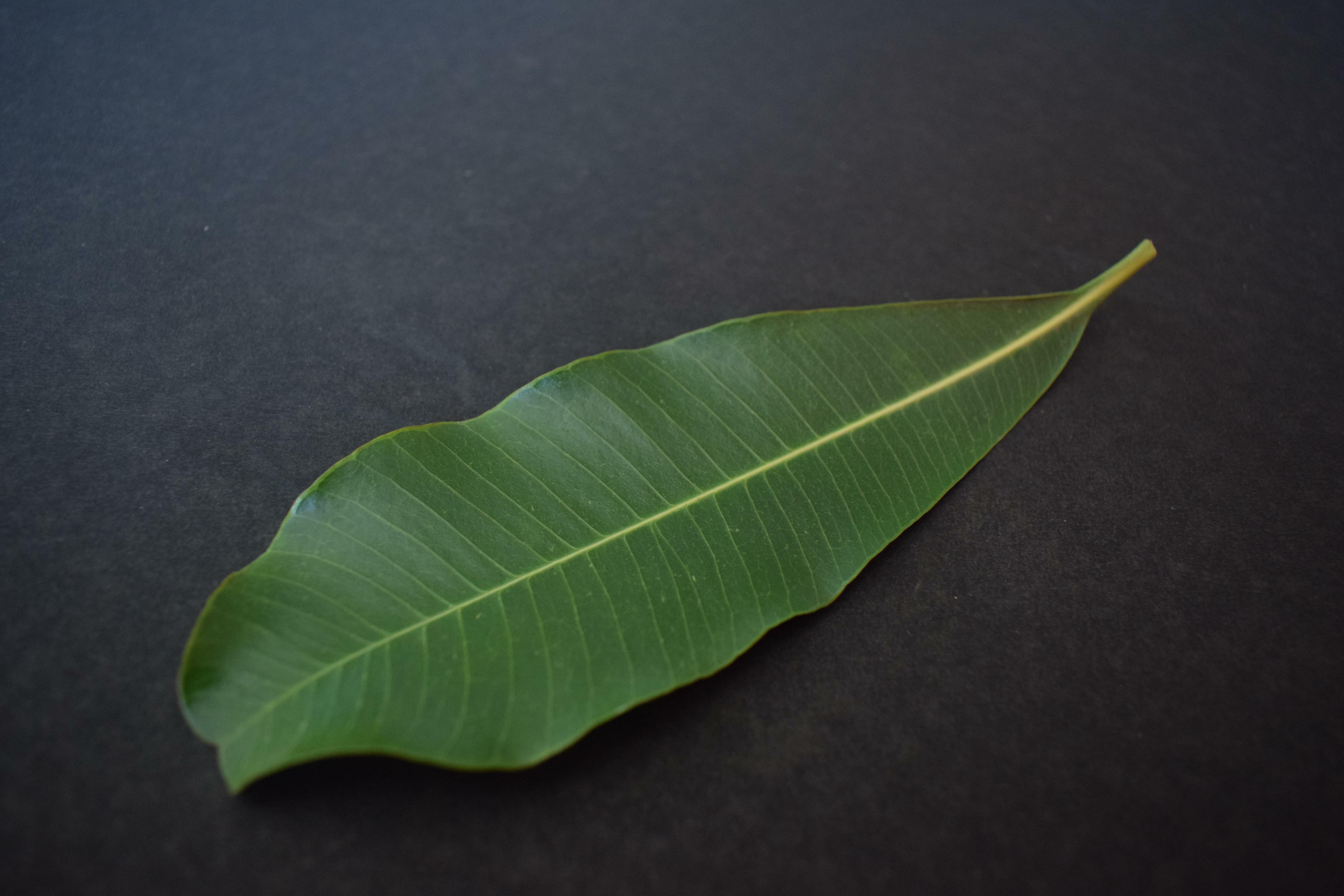}
        \includegraphics[width=0.28\textwidth]{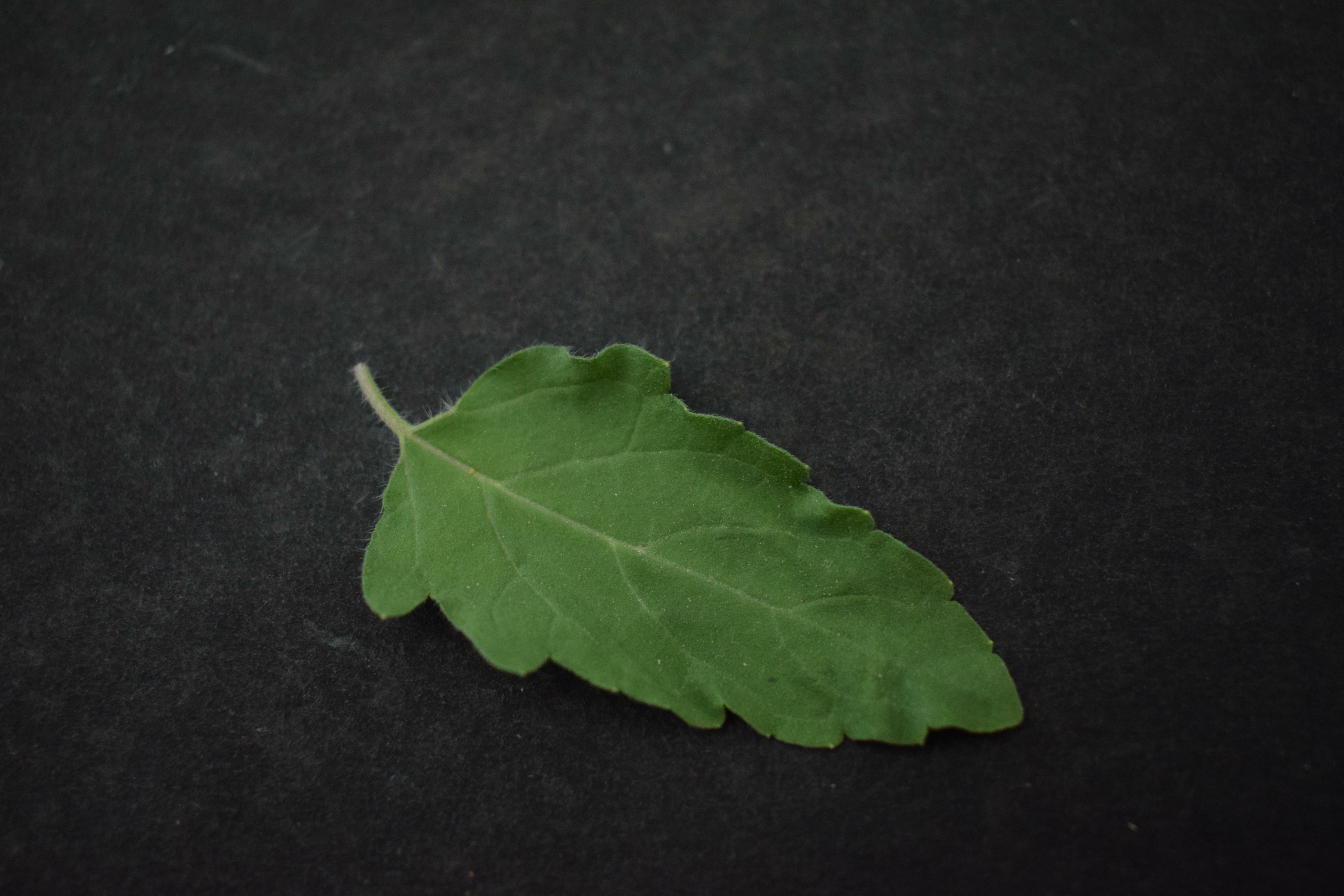}
        \includegraphics[width=0.28\textwidth]{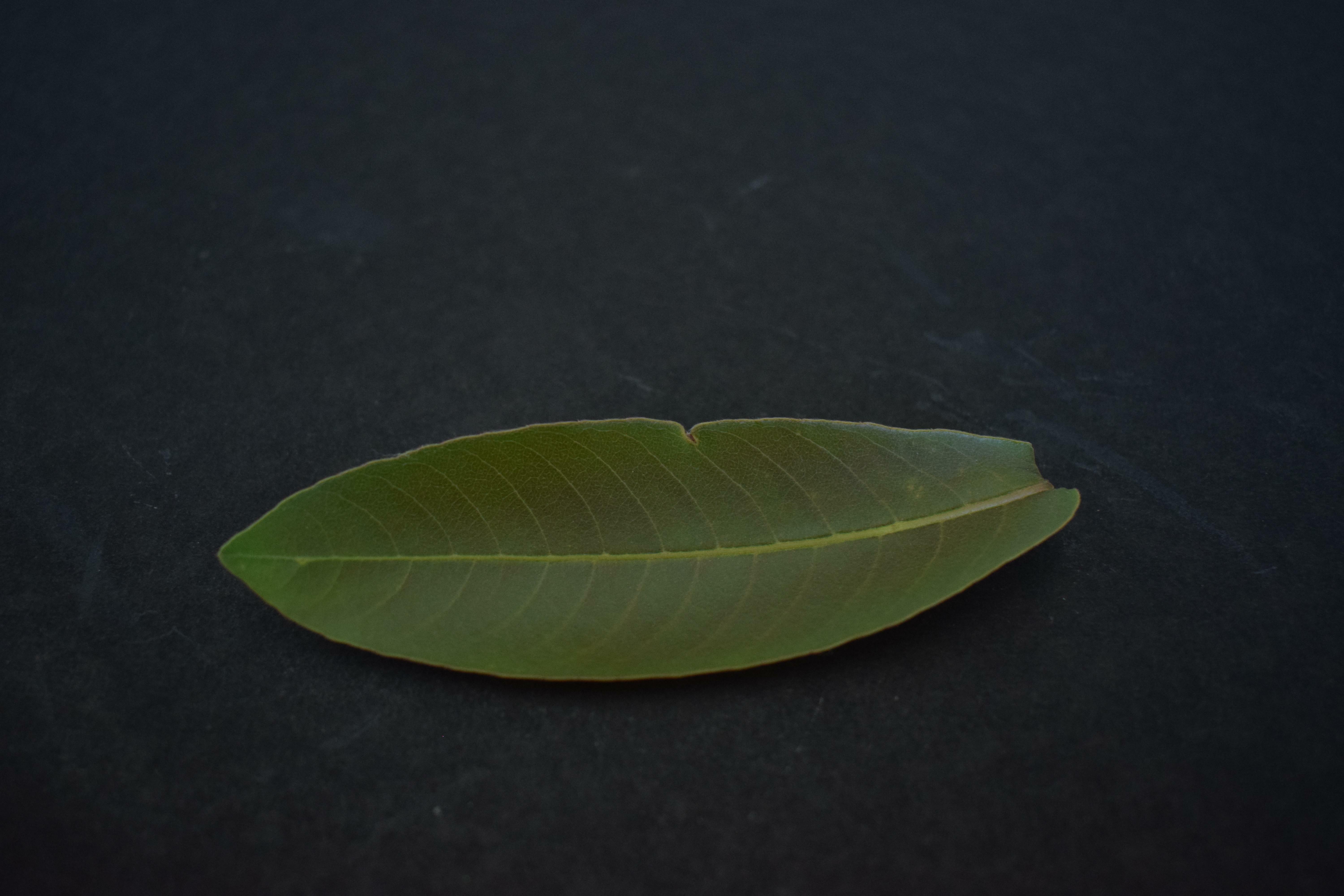}
        \caption{Plant leaves}
        \label{fig:plant_leaves}
             \end{subfigure}       

                \begin{subfigure}[b]{0.35\textwidth}
         \centering
        \includegraphics[width=0.28\textwidth]{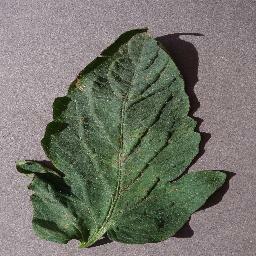}
        \includegraphics[width=0.28\textwidth]{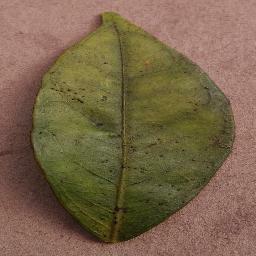}
        \includegraphics[width=0.28\textwidth]{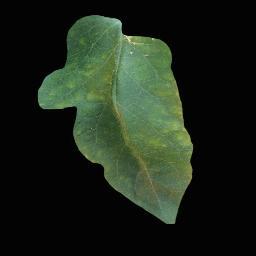}
        \caption{Plant village}
        \label{fig:plant_village}
        
     \end{subfigure}       

\caption{Samples images of the nine datasets used in the experiments.}
\label{fig:images}
\end{figure}

\begin{table*}[htpb]
\centering
\resizebox{0.7\textwidth}{!}{%
\begin{tabular}{@{}ccccccc@{}}
\toprule
\textbf{} & \textbf{Version} & \textbf{No. labels} & \textbf{Train size} & \textbf{Valid. size} & \textbf{Test size} & \textbf{Ref.} \\ \midrule
\textbf{Caltech birds 2011} & 0.1.1 & 200 & 5,395 & 599 & 5,794 & \cite{WelinderEtal2010} \\ \midrule
\textbf{Cars 196} & 2.0.0 & 196 & 7,330 & 814 & 8,041 & \cite{KrauseStarkDengFei-Fei_3DRR2013} \\ \midrule
\textbf{Cassava} & 0.1.0 & 5 & 5,656 & 1,889 & 1,885 & \cite{mwebaze2019icassava}\\ \midrule
\textbf{Cats vs dogs} & 4.0.0 & 2 & 16,283 & 2,327 & 4,652 & \cite{asirra2007elson} \\ \midrule
\textbf{Citrus leaves} & 0.1.2 & 4 & 416 & 59 & 119 & \cite{rauf2019citrus} \\ \midrule
\textbf{Deep weeds} & 3.0.0 & 9 & 12,256 & 1,751 & 3,502 & \cite{DeepWeeds2019} \\ \midrule
\textbf{Malaria} & 1.0.0 & 2 & 19,291 & 2,755 & 5,512 & \cite{rajaraman2018pre} \\ \midrule
\textbf{Oxford flowers 102} & 2.1.1 & 102 & 1,020 & 1,020 & 6,149 & \cite{Nilsback08} \\ \midrule
\textbf{Plant leaves} & 0.1.0 & 22 & 3,151 & 451 & 900 & \cite{plant2019siddhart} \\ \midrule
\textbf{Plant village} & 1.0.2 & 38 & 38,012 & 5,430 & 10,861 & \cite{DBLP:journals/corr/HughesS15} \\ \bottomrule
\end{tabular}%
}
\caption{Datasets used in the experimentation, showing the version number, number of labels and the data splits size. See Figure \ref{fig:images} for example images.}
\label{tab:datasets_data}
\end{table*}

    The concern of this research is to present a powerful method for improving CNNs independent of the applied domain. Therefore, a wide set of 10 heterogeneous datasets has been selected to validate our methods. These are summarised as follows:  

    \begin{enumerate}[a)]
    
        \item \textit{Caltech birds2011}~\cite{WelinderEtal2010}: This dataset contains images from mostly north American bird species in different poses and perspectives, usually at the centre of the image. This set contains 11.788 instances distributed in 200 classes.
        
        \item \textit{Cars 196}~\cite{KrauseStarkDengFei-Fei_3DRR2013}: This dataset contains car images without context, where the vehicle is the focus with varying perspectives. There are 16.185 instances evenly distributed on 196 car types.  
        
        \item \textit{Cassava}~\cite{mwebaze2019icassava}: A collection of 9430 images of the cassava plant, including healthy and four types of disease, with high unbalance between classes.

        \item \textit{Cats vs dogs}~\cite{asirra2007elson}: This dataset is a binary classification problem of cat and dog images with context, in some instances sharing space with a human, out of focus or with obstructions. This dataset contains 23.262 evenly distributed images among both classes. 
        
        \item \textit{Citrus leaves}~\cite{rauf2019citrus}: This dataset has a low number of images from healthy and unhealthy citrus fruit leaves. This dataset has 759 instances distributed in 4 classes. 

        \item \textit{Deep weeds}~\cite{DeepWeeds2019}: This dataset contains weed images from the grasslands of Queensland, Australia, with the aim of telling apart weeds from grass, and identifying the type of weed. It contains 17,509 instances evenly distributed across 9 classes. In the original article, Inception v3 and ResNet-50 are used with outstanding results. 

        \item \textit{Malaria}~\cite{rajaraman2018pre}: This dataset is a repository of 27.558 segmented cell images, with an equal amount of instances of parasitized cells and uninfected cells, with the aim of improving diagnostic accuracy of malaria cases. The original article achieves powerful results with a ResNet-50. 
                
        \item \textit{Oxford flowers102}~\cite{Nilsback08}: This dataset has flower images of different species, focused on the flower with limited context. It contains 8.189 instances distributed across 102 classes, with 40 to 258 images per label. 
        
        \item \textit{Plant leaves}~\cite{plant2019siddhart}: This dataset contains plant images without context of different leaves species and health condition. This requires detection of both: plant type and plant health. There are 4.502 images of 22 classes total, evenly distributed.  

        \item \textit{Plant village}~\cite{DBLP:journals/corr/HughesS15}: This dataset contains a large number of leaf images, focused and without context. This dataset contains 54.303 instances of leaves, distributed evenly in 38 classes.

    \end{enumerate}
    
    Table~\ref{tab:datasets_data} shows a summary of all these datasets, including version, number of labels and size of every split. The default partitions were kept. In those cases where only one (training) or two (training and test) splits were defined, a new partitioning was made to allocate 70\% instances for training, 10\% for validation and 20\% for testing. Figure \ref{fig:images} shows example images for all datasets.

    \subsection{Image preprocessing}
    
    In order to modify every image as little as possible, only two preprocessing steps were followed before passing through the CNN architecture. First, since the datasets provide pictures with different sizes, every image was resized to a 224x224 format. Then, every pixel value was normalised from [0,255] to [0, 1].
    
    \subsection{CNN architectures}
    \label{sec:cnns}
    Five different architectures were tested in this research:
    
            
            \begin{itemize}
                \item \textit{VGG19}~\cite{simonyan2014very}: The VGG16 and VGG19 architectures are composed by a feed-forward set of units, containing 2 to 4 convolutional layers, an activation and a pooling layer. Out of all the architectures here, it is the most straightforward, using no additional forward connections or auxiliary outputs. This network has a very large amount of parameters to adjust.
                \item \textit{DenseNet}~\cite{Huang2017Jul}: DenseNet, instead of relying on adding more units to its design, it strengthens the number of connections between layers. The main idea behind this design is to forward connect every unit. Each unit has again several convolutional layers, a batch normalisation layer (for regularisation), activation and pooling. 
                \item InceptionV3~\cite{Szegedy2015Dec}: The following architectures stem from the same family of architectures. This architecture factorises convolutions into simpler operations, for example a 5x5 convolution into two 3x3 convolutions. These factorisations are separated in different pipelines inside a unit and several units are concatenated to achieve the end result.
                \item Xception~\cite{Szegedy2015Dec}: Xception proposes a depth wise separable convolution which is an extension of the factorisation proposed on InceptionV3. This convolutional operation involves two steps: first, an spatial convolution, transforming the channel, followed by a pointwise convolution, a 1x1 filter over the channels. Xception also incorporates feedforward residual connections, similar to a ResNet~\cite{He2015Dec}, connecting the last layer to the next one via a single convolution and an addition operation.
                \item InceptionResNetV2~\cite{Szegedy2016Feb}: This architecture combines the ideas from InceptionV3 and ResNet. By stacking several Inception units, and connecting them with residual feedforward addition operations, the capabilities of the network are enhanced.
            \end{itemize}
            
            
            All these architectures were initialised with transferred weights from the ImageNet domain, a percent of which are frozen and connected to a dense layer with 50\% dropout. The optimisation algorithm chosen is Adam, while any other remaining hyper-parameter has been tuned empirically. The end result of the networks is an output vector of probabilities from the last dense layer (with softmax activation). These probabilities have values between 0 and 1, summing 1 between them, and the argmax is selected as the true label for classification purposes.
        
    \subsection{CNN models execution}
        
        Every CNN model listed in Section~\ref{sec:cnns} was executed five times with the Keras library for Python~\cite{chollet2015keras} with the TensorFlow~\cite{tensorflow2015-whitepaper} backend for each dataset, starting from pre-trained Keras models. All CNN relevant hyperparameters are summarized in Table \ref{tab:hyperparameters}. Then, all CNNs were fitted using the categorical cross-entropy as loss function, the Adam optimiser with gradient norm scaling if the vector exceeds $1.0$, a learning rate of $0.0001$, and fixing a maximum of 1,000 epochs. An early stopping criteria was also set to stop the training when the validation accuracy did not improve in the last 25 epochs more than a threshold $\theta = 0.0001$. Additionally, the training of every model was performed using a batch size of 64 examples. All datasets were streamlined by the TensorFlow dataset tools. 
        
\begin{table}[htpb]
\centering
\resizebox{\linewidth}{!}{%
\begin{tabularx}{1\linewidth}{@{}lY@{}}
\toprule
\multicolumn{1}{l}{\textbf{First layers frozen (\%)}} &           \\ \midrule
\textit{VGG19}                                        & 0\%       \\
\textit{InceptionResNetV2}                            & 25\%      \\
\textit{InceptionV3}                                  & 10\%      \\
\textit{DenseNet201}                                  & 25\%      \\
\textit{XceptionV1}                                   & 25\%      \\
&\\
\multicolumn{1}{l}{\textbf{Adam optimizer}}           &           \\ \midrule
\textit{$\alpha$ Learning Rate}                       & 1e-4      \\
\textit{$\beta_1$ Momentum 1}                         & 0.9       \\
\textit{$\beta_2$ Momentum 2}                         & 0.999     \\
\textit{$\epsilon$ Stability constant}                & 1e-7      \\
\textit{Gradient normalization}                       & 1         \\
&\\
\multicolumn{1}{l}{\textbf{Feed-forward classifier}}  & \textbf{} \\ \midrule
\textit{Global average pooling}                       &           \\
\textit{Neurons}                                      & 2048      \\
\textit{Dropout}                                      & 50\%      \\
\textit{Activation}                                   & ReLU      \\
&\\
\multicolumn{1}{l}{\textbf{Training methodology}}     &           \\ \midrule
\textit{Maximum epochs}                               & 1000      \\
\textit{Early stopping patience}                      & 25        \\
\textit{$\theta$ Early stopping threshold}            & 1e-4      \\
\textit{Batch size}                                   & 64        \\
\textit{Image size (NxN)}                             & N=224     \\ \bottomrule
\end{tabularx}%
}
\caption{Summary of hyper-parameters relevant to CNN networks transfer learning, optimization, classification and training.}
\label{tab:hyperparameters}
\end{table}
    \subsection{ML models hyperparameters}
    
        For the execution of the classification algorithms listed in Section~\ref{sec:statistical_indicators}, the scikit-learn~\cite{scikit-learn} was used, using default parameters for all of them. In the case of Random Forest, 500 internal trees were employed. Due to the randomness of the Random Forest classifier when assigning samples to estimators and selection of features during the definition of new branches in the trees, it was calculated the average of 10 different executions.

    \subsection{Execution environment \& Github repository}
         All executions have been run in a machine with a 48GB Nvidia Quadro RTX 8000, an Intel(R) Xeon(R) Bronze 3206R CPU $@$ 1.90GHz and 256GB RAM.\\

        
        The code developed for this article is publicly available and can be found at: \url{https://github.com/jahuerta92/cnn-prob-ensemble}.

\begin{figure*}  
	\centering
	\includegraphics[width=1\textwidth]{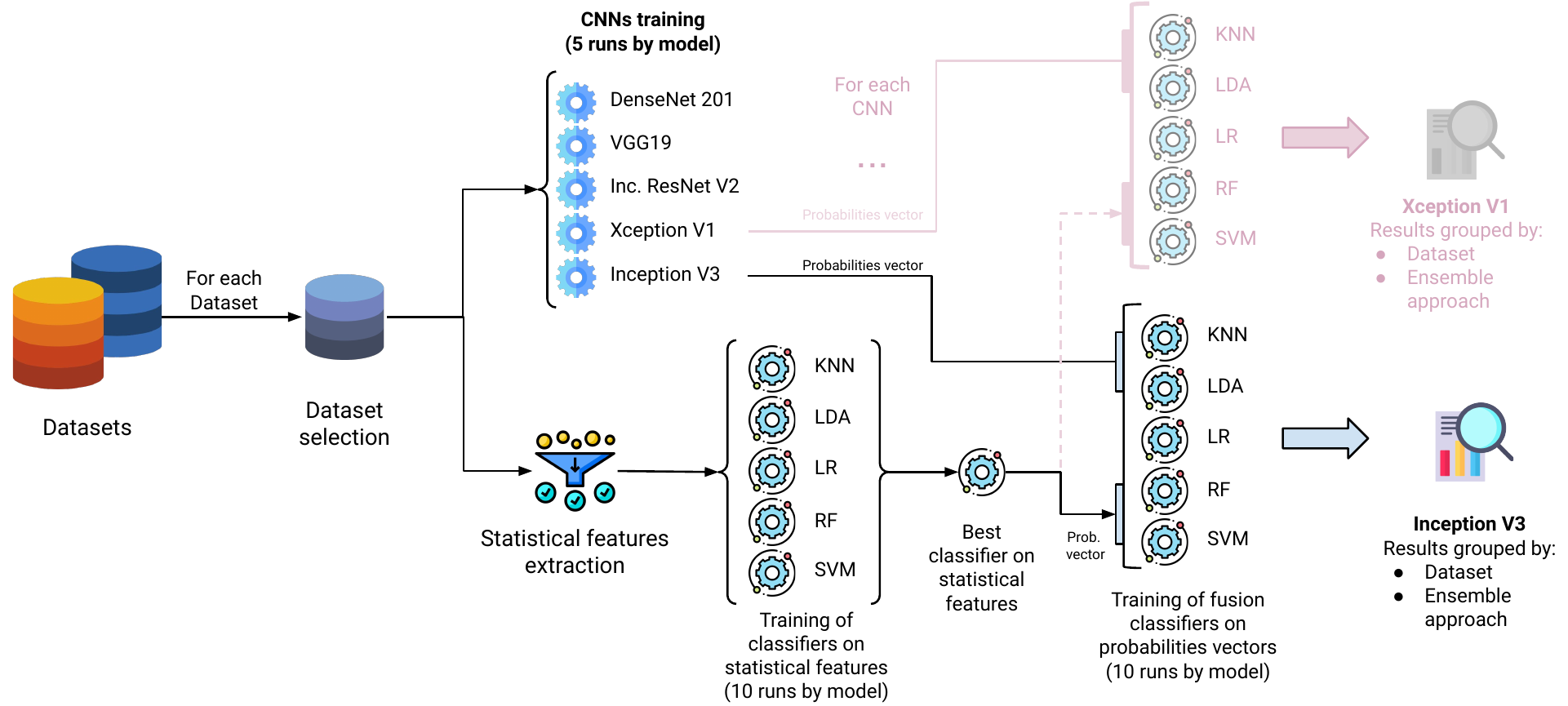}
	\caption{This flowchart shows how the experiments are conducted. For each dataset considered, 5 different CNN architectures are trained and the set of statistical features is extracted. With this last set, different classical machine learning classifiers are trained, selecting the one which performed better. Then, for each CNN model trained, it is evaluated how it can be combined with statistical features based classification algorithm, using again different machine learning classification algorithms. Finally, the results show how, for each dataset and each specific CNN architecture, an ensemble can be built in order to improve the results of the CNN base architecture in isolation.}
	\label{fig:7_experimental_setup}
\end{figure*}

\begin{table*}[htpb]
\centering
\resizebox{\textwidth}{!}{%
\begin{tabular}{@{}cccccccccccc@{}}
\toprule
\multirow{2}{*}{\textbf{CNN model}} &
  \multirow{2}{*}{\textbf{Measure}} &
  \multicolumn{10}{c}{\textbf{Dataset}} \\ \cmidrule(l){3-12} 
 &
   &
  \textbf{\begin{tabular}[c]{@{}c@{}}Caltech\\ birds\\ 2011\end{tabular}} &
  \textbf{Cars196} &
  \textbf{Cassava} &
  \textbf{\begin{tabular}[c]{@{}c@{}}Cats vs\\ dogs\end{tabular}} &
  \textbf{\begin{tabular}[c]{@{}c@{}}Citrus\\ leaves\end{tabular}} &
  \textbf{\begin{tabular}[c]{@{}c@{}}Deep\\ weeds\end{tabular}} &
  \textbf{Malaria} &
  \textbf{\begin{tabular}[c]{@{}c@{}}Oxford\\ flowers\\ 102\end{tabular}} &
  \textbf{\begin{tabular}[c]{@{}c@{}}Plant\\ leaves\end{tabular}} &
  \textbf{\begin{tabular}[c]{@{}c@{}}Plant\\ village\end{tabular}} \\ \midrule
  
\multirow{3}{*}{\textbf{VGG19}} &
  \textbf{Max.} &
  0.0\% &
  0.01\% &
  78.8\% &
  98.24\% &
  96.15\% &
  92.15\% &
  96.59\% &
  32.47\% &
  94.49\% &
  99.24\% \\
 &
  \textbf{Mean} &
  0.0\% &
  0.01\% &
  76.55\% &
  97.93\% &
  \textbf{92.81\%} &
  90.73\% &
  82.0\% &
  16.03\% &
  93.44\% &
  99.18\% \\
 &
  \textbf{Std.} &
  0.0\% &
  0.0\% &
  1.96\% &
  0.35\% &
  2.96\% &
  1.15\% &
  31.99\% &
  15.03\% &
  1.03\% &
  0.12\% \\ \midrule
\multirow{3}{*}{\textbf{\begin{tabular}[c]{@{}c@{}}DenseNet \\ 201\end{tabular}}} &
  \textbf{Max.} &
  73.08\% &
  82.92\% &
  82.52\% &
  99.34\% &
  94.28\% &
  93.97\% &
  97.33\% &
  85.33\% &
  96.83\% &
  99.66\% \\
 &
  \textbf{Mean} &
  \textbf{72.55\%} &
  \textbf{82.28\%} &
  \textbf{81.71\%} &
  \textbf{99.17\%} &
  91.96\% &
  \textbf{93.7\%} &
  \textbf{97.16\%} &
  \textbf{85.04\%} &
  \textbf{96.14\%} &
  \textbf{99.58\%} \\
 &
  \textbf{Std.} &
  0.55\% &
  0.46\% &
  0.92\% &
  0.17\% &
  1.67\% &
  0.28\% &
  0.15\% &
  0.19\% &
  0.49\% &
  0.08\% \\ \midrule
  
\multirow{3}{*}{\textbf{\begin{tabular}[c]{@{}c@{}}Inception\\ V3\end{tabular}}} &
  \textbf{Max.} &
  61.78\% &
  72.54\% &
  76.68\% &
  99.0\% &
  89.05\% &
  91.26\% &
  97.23\% &
  69.67\% &
  95.24\% &
  99.44\% \\
 &
  \textbf{Mean} &
  61.49\% &
  68.39\% &
  75.05\% &
  98.78\% &
  86.7\% &
  90.06\% &
  96.95\% &
  68.6\% &
  94.55\% &
  99.32\% \\
 &
  \textbf{Std.} &
  0.36\% &
  4.38\% &
  1.24\% &
  0.16\% &
  2.6\% &
  1.15\% &
  0.2\% &
  0.73\% &
  0.58\% &
  0.07\% \\ \midrule

\multirow{3}{*}{\textbf{\begin{tabular}[c]{@{}c@{}}Xception\\ V1\end{tabular}}} &
  \textbf{Max.} &
  62.78\% &
  71.32\% &
  76.3\% &
  98.97\% &
  91.65\% &
  87.43\% &
  97.08\% &
  78.49\% &
  95.17\% &
  99.55\% \\
 &
  \textbf{Mean} &
  62.08\% &
  67.97\% &
  75.95\% &
  98.82\% &
  89.25\% &
  87.13\% &
  96.87\% &
  77.82\% &
  94.63\% &
  99.41\% \\
 &
  \textbf{Std.} &
  0.59\% &
  2.31\% &
  0.43\% &
  0.14\% &
  1.78\% &
  0.43\% &
  0.22\% &
  0.54\% &
  0.38\% &
  0.08\% \\ \midrule
  
  \multirow{3}{*}{\textbf{\begin{tabular}[c]{@{}c@{}}Inception\\ ResNet\\ V2\end{tabular}}} &
  \textbf{Max.} &
  68.27\% &
  77.21\% &
  78.4\% &
  99.2\% &
  92.49\% &
  91.61\% &
  97.28\% &
  79.21\% &
  96.37\% &
  99.58\% \\
 &
  \textbf{Mean} &
  67.2\% &
  75.26\% &
  77.43\% &
  99.04\% &
  89.74\% &
  90.6\% &
  97.11\% &
  77.94\% &
  95.81\% &
  99.44\% \\
 &
  \textbf{Std.} &
  0.79\% &
  1.42\% &
  0.7\% &
  0.1\% &
  1.76\% &
  0.72\% &
  0.22\% &
  0.94\% &
  0.42\% &
  0.13\% \\ \bottomrule
\end{tabular}%
}
\caption{Averaged results of the five CNN architectures evaluated in 10 different datasets. All values indicate the macro average weighted precision. The maximum row refers to the best execution according to the precision obtained in the validation set.}
\label{tab:cnn_results}
\end{table*}

\section{Experimentation}
\label{sec:experimentation}

This section describes all the experiments performed and the results obtained. First, the performance of the base five CNN architectures is evaluated, showing how each of them performs in each dataset. Then, the use of statistical features is assessed in all datasets using different machine learning classifiers to later evaluate the performance of the proposed fusion approach in comparison with the performance of the CNN architectures. Then, an ablation study helps to analyse the individual contribution of each statistical feature considered. Finally, a brief analysis of the time performance of the approach is provided.

In order to better describe how all the experimentation was conducted, Figure~\ref{fig:7_experimental_setup} shows an overview of the different steps followed. The fusion approach was evaluated independently in 10 different datasets for 5 different CNN architectures. In case of the classification model based on statistical information, different machine learning methods with default hyperparameters were tested, selecting the one which performed better. Finally, for each dataset and architecture, different classifiers were again tested, with the goal of building the best possible combination between the probability vector obtained from the CNN and vector obtained from the best classifier trained on the statistical information.

\begin{table*}[htpb]
\centering
\resizebox{\textwidth}{!}{%
\begin{tabular}{@{}ccccccccccc@{}}
\toprule
\multirow{2}{*}{\textbf{Classifier}} & \multicolumn{10}{c}{\textbf{Dataset}} \\ \cmidrule(l){2-11} 
 & \textbf{\begin{tabular}[c]{@{}c@{}}Caltech\\ birds\\ 2011\end{tabular}} & \textbf{Cars196} & \textbf{Cassava} & \textbf{\begin{tabular}[c]{@{}c@{}}Cats vs\\ dogs\end{tabular}} & \textbf{\begin{tabular}[c]{@{}c@{}}Citrus\\ leaves\end{tabular}} & \textbf{\begin{tabular}[c]{@{}c@{}}Deep\\ weeds\end{tabular}} & \textbf{Malaria} & \textbf{\begin{tabular}[c]{@{}c@{}}Oxford\\ flowers\\ 102\end{tabular}} & \textbf{\begin{tabular}[c]{@{}c@{}}Plant\\ leaves\end{tabular}} & \textbf{\begin{tabular}[c]{@{}c@{}}Plant\\ village\end{tabular}} \\ \midrule
\textbf{Logistic Regression} & 2.38\% & 1.89\% & 56.16\% & 62.34\% & 88.74\% & 54.05\% & 84.95\% & 22.09\% & 66.05\% & 63.0\% \\
\textbf{LD Analysis} & \textbf{3.44\%} & \textbf{3.6\%} & 64.32\% & \textbf{63.51\%} & \textbf{89.41\%} & 61.86\% & 90.17\% & \textbf{33.19\%} & \textbf{81.98\%} & 76.27\% \\
\textbf{KNeighbors} & 2.58\% & 2.56\% & 55.94\% & 58.87\% & 86.7\% & 65.35\% & 78.42\% & 24.71\% & 72.5\% & 77.79\% \\
\textbf{SVM-rbf} & 2.74\% & 1.63\% & \textbf{66.51\%} & 62.98\% & 87.57\% & 64.44\% & 89.1\% & 22.41\% & 70.27\% & 74.57\% \\
\textbf{SVM-sigmoid} & 0.07\% & 0.01\% & 35.79\% & 58.81\% & 42.32\% & 29.61\% & 49.62\% & 13.99\% & 6.83\% & 20.09\% \\
\textbf{Random Forest} & 3.1\% & 3.59\% & 60.86\% & 63.43\% & 85.62\% & \textbf{68.72\%} & \textbf{95.84\%} & 23.09\% & 76.62\% & \textbf{79.86\%} \\ \bottomrule
\end{tabular}%
}
\caption{Results of different classifiers trained with the statistical features extracted from every image. All values indicate the macro average weighted precision.}
\label{tab:results_combination}
\end{table*}

\subsection{CNN models performance}
\label{sec:cnn_results}

The first step is to train and evaluate several times each CNN architecture defined in Section~\ref{sec:cnns}. This serves as a baseline for the performance of these models to later quantify the improvement achieved by the addition of statistical manually-extracted features. The results, in terms of macro average weighted precision, are shown in Table~\ref{tab:cnn_results}. For each dataset and each CNN architecture, the table shows the average and standard deviation of macro average weighted precision in the test set for each execution and the best result according to the validation set. DenseNet presents the best results among the different datasets used, reaching the best position in 9 of 10 datasets. One of the main characteristics of this model lies in that it has an important number or parameters (20,242,984) and a topological depth of 201 elements (for the DenseNet 201 version considered in this research). Despite not being the most complex architecture, DenseNet shows an excellent performance among datasets. Inception ResNet V2, although it is a more complex architecture, with 55,873,736 parameters a topology composed of 572 elements shows slightly worse results. 

VGG-19, although also produces the best result in one domain (Citrus leaves) and also excellent results in other four domains (Cats vs dogs, Deep weeds, Plant leaves and Plant Village), encountered severe difficulties in training an accurate model in 3 domains: Cars 196, Caltech birds2011, and Oxford flowers102. This effect coincides with the fact that these datasets have a large number of labels, more than 100 in the three cases, leading to conclude that VGG19 does not allow to retrieve enough discriminating information to distinguish between such a large number of labels. It is worth mentioning that VGG tends to erroneously classify all samples to the same label during the training phase. Additionally, this low precision could also be caused due to the weights of the pre-trained model or the necessity of modifying certain hyperparameters. In order to avoid any type of biased decision, we run all models with default parameters.

In almost all datasets there is room for improvement. However, in the case of Plant village and Cats vs dogs, the results surpassed 99\% precision, so it is not expected to increase this value with the inclusion of statistical information. However, we decided to include these datasets in order to verify that the use of additional features and a second classification step does not lead to a counterproductive approach.

\subsection{Image classification based on statistical indicators}
\label{sec:statistical_results}

The second step of the experimentation followed involves evaluating the capacity of the statistical features manually extracted to reveal important patterns that can help to differentiate between labels. This new representation, a vector of 54 values extracted from the image (as shown in Table~\ref{tab:features}), was tested using different machine learning classifiers, obtaining the results shown in Table~\ref{tab:results_combination}. As in the case of the evaluation of the base CNN architectures, and given the huge imbalance between classes, the macro average weighted precision was used. 

As can be seen in the table, LDA (Linear Discriminant Analysis) and Random Forests are the stronger models to classify images according to statistical features. It is remarkable how the statistical features extracted retain a different degree of discriminating information depending on the domain. While on datasets such as Citrus leaves or Malaria, the result is very close to the one obtained with a CNN network, in case of Cars196 and Caltech birds2011, the classifiers reach low precision rates. 

In Citrus leaves and Malaria, the solely use of these statistical indicators allows to reach close or more than 89\% precision. This is a very remarkable result. The CNN architectures trained in these domains improved this figure only by 2\% or 3\%, but making use of extraordinary bigger computation and time resources. Moreover, the use of statistical indicators provide a useful instrument in specific domains, as we can know exactly what every feature means, instead of the complex and convoluted features extracted within the CNN architecture. The use of statistical features manually extracted together with a rule-base classifier, for instance, Random Forests, provides an explainable classification path~\cite{paco2020xai}, rather than a black box~\cite{2021_WACV_XAI_Alfonso}, as it is the case of a Convolutional Neural Network. In the next subsection we evaluate if the combination of classification probabilities obtained from a CNN architecture and the help of a classification model based on statistical indicators leads to a stronger classification approach.

\begin{table*}[htpb]
\centering
\resizebox{\textwidth}{!}{%
\begin{tabular}{@{}ccc|ccccccc@{}}
\toprule
\multirow{2}{*}{\textbf{Dataset}} & \multirow{2}{*}{\textbf{\begin{tabular}[c]{@{}c@{}}CNN\\ architecture\end{tabular}}} & \multirow{2}{*}{\textbf{\begin{tabular}[c]{@{}c@{}}Base CNN\\ (avg. weighted\\ precision)\end{tabular}}} & \multicolumn{7}{c}{\textbf{Ensemble (avg. weighted precision)}} \\ \cmidrule(l){4-10} 
 &  &  & \textbf{Avg.} & \textbf{KNN} & \textbf{LDA} & \textbf{LR} & \textbf{RF} & \textbf{\begin{tabular}[c]{@{}c@{}}SVM\\ rbf\end{tabular}} & \textbf{\begin{tabular}[c]{@{}c@{}}SVG\\ sig\end{tabular}} \\ \midrule
\multirow{5}{*}{\textbf{\begin{tabular}[c]{@{}c@{}}Caltech\\ birds\\ 2011\end{tabular}}} 
 & \textbf{VGG19} & 0.0±0.0\% & 3.45\% & 3.17\% & \textbf{4.15\%} & 3.09\% & 3.69±0.14\% & 3.38\% & 3.65\% \\
 & \textbf{DenseNet 201} & 72.55±0.55\% & 70.09\% & 72.65\% & 74.31\% & 72.85\% & \textbf{78.49±0.33\%} & 72.58\% & 72.89\% \\
 & \textbf{Inception V3} & 61.49±0.36\% & 59.66\% & 61.49\% & 64.57\% & 61.72\% & \textbf{68.42±0.62\%} & 61.94\% & 61.69\% \\
 & \textbf{Xception V1} & 62.08±0.59\% & 60.78\% & 62.44\% & 63.99\% & 62.5\% & \textbf{64.83±0.63\%} & 62.61\% & 62.47\% \\
 & \textbf{Inc. ResNet V2} & 67.2±0.79\% & 65.95\% & 66.87\% & 67.21\% & 67.05\% & \textbf{69.72±0.42\%} & 66.94\% & 66.93\% \\ \midrule

\multirow{5}{*}{\textbf{\begin{tabular}[c]{@{}c@{}}Cars 196\end{tabular}}} 
 & \textbf{VGG19} & 0.01±0.0\% & 3.6\% & 3.39\% & 3.73\% & 2.9\% & 4.35±0.19\% & 4.52\% & \textbf{4.54\%} \\
 & \textbf{DenseNet 201} & 82.28±0.46\% & 81.02\% & 81.94\% & 81.72\% & 81.81\% & \textbf{84.66±0.15\%} & 82.2\% & 81.96\% \\
 & \textbf{Inception V3} & 68.39±4.38\% & 71.7\% & 72.65\% & 72.97\% & 72.53\% & \textbf{75.94±0.24\%} & 73.0\% & 72.78\% \\
 & \textbf{Xception V1} & 67.97±2.31\% & 70.49\% & 71.44\% & 71.35\% & 71.3\% & \textbf{73.48±0.24\%} & 71.52\% & 71.34\% \\ 
 & \textbf{Inc. ResNet V2} & 75.26±1.42\% & 74.46\% & 75.39\% & 73.9\% & 75.18\% & \textbf{77.52±0.23\%} & 76.44\% & 75.3\% \\ \midrule

\multirow{5}{*}{\textbf{Cassava}} 
 & \textbf{VGG19} & 76.55±1.96\% & 83.93\% & \textbf{84.12\%} & 83.96\% & 84.09\% & 84.0±0.09\% & 83.96\% & 84.03\% \\
 & \textbf{DenseNet 201} & 81.71±0.92\% & 86.62\% & 86.85\% & 86.92\% & 86.99\% & 86.97±0.09\% & 86.76\% & \textbf{87.06\%} \\
  & \textbf{Inception V3} & 75.05±1.24\% & \textbf{81.58\%} & 81.28\% & 81.38\% & 81.29\% & 81.41±0.08\% & 81.33\% & 81.56\% \\
 & \textbf{Xception V1} & 75.95±0.43\% & \textbf{81.18\%} & 80.66\% & 80.63\% & 80.61\% & 80.3±0.12\% & 80.43\% & 80.42\% \\ 
 & \textbf{Inc. ResNet V2} & 77.43±0.7\% & \textbf{82.83\%} & 82.44\% & 82.63\% & 82.56\% & 82.42±0.13\% & 82.46\% & 82.49\% \\ \midrule

\multirow{5}{*}{\textbf{\begin{tabular}[c]{@{}c@{}}Cats vs\\ dogs\end{tabular}}} 
 & \textbf{VGG19} & 97.93±0.35\% & 98.26\% & 98.28\% & 98.24\% & 98.24\% & 98.14±0.03\% & \textbf{98.3\%} & 98.26\% \\
 & \textbf{DenseNet 201} & 99.17±0.17\% & 99.29\% & 99.33\% & 99.33\% & 99.33\% & \textbf{99.36±0.0\%} & \textbf{99.36\%} & 99.33\% \\
  & \textbf{Inception V3} & 98.78±0.16\% & \textbf{98.84\%} & 98.8\% & 98.8\% & 98.8\% & 98.82±0.0\% & 98.82\% & 98.8\% \\
 & \textbf{Xception V1} & 98.82±0.14\% & 98.88\% & \textbf{98.97\%} & \textbf{98.97\%} & \textbf{98.97\%} & \textbf{98.97±0.0\%} & \textbf{98.97\%} & \textbf{98.97\%} \\ 
 & \textbf{Inc. ResNet V2} & 99.04±0.1\% & 98.97\% & 99.08\% & 98.97\% & 98.97\% & \textbf{99.09±0.01\%} & 98.99\% & 98.95\% \\ \midrule

\multirow{5}{*}{\textbf{\begin{tabular}[c]{@{}c@{}}Citrus\\ leaves\end{tabular}}} 
 & \textbf{VGG19} & 92.81±2.96\% & \textbf{95.34\%} & 92.61\% & 91.68\% & 92.61\% & 92.54±0.45\% & 91.68\% & 93.38\% \\
 & \textbf{DenseNet 201} & 91.96±1.67\% & \textbf{94.42\%} & 91.0\% & 91.0\% & 91.82\% & 91.0±0.0\% & 91.0\% & 91.0\% \\
  & \textbf{Inception V3} & 86.7±2.6\% & \textbf{92.73\%} & 88.66\% & 86.69\% & 89.33\% & 87.35±1.08\% & 88.18\% & 87.67\% \\
 & \textbf{Xception V1} & 89.25±1.78\% & \textbf{94.23\%} & 89.12\% & 87.09\% & 90.1\% & 87.98±0.5\% & 87.98\% & 88.35\% \\ 
 & \textbf{Inc. ResNet V2} & 89.74±1.76\% & \textbf{92.47\%} & 89.48\% & 89.48\% & 90.18\% & 89.48±0.59\% & 89.42\% & 90.17\% \\ \midrule

\multirow{5}{*}{\textbf{\begin{tabular}[c]{@{}c@{}}Deep\\ weeds\end{tabular}}} 
 & \textbf{VGG19} & 90.73±1.15\% & 91.41\% & \textbf{91.43\%} & 91.13\% & 91.39\% & 84.21±1.31\% & 91.18\% & 91.07\% \\
 & \textbf{DenseNet 201} & 93.7±0.28\% & 94.54\% & \textbf{94.67\%} & 94.64\% & 94.59\% & 88.88±2.72\% & 94.61\% & 94.66\% \\
  & \textbf{Inception V3} & 90.06±1.15\% & 90.96\% & 90.92\% & \textbf{91.05\%} & 91.02\% & 85.38±1.19\% & 90.95\% & 90.9\% \\
 & \textbf{Xception V1} & 87.13±0.43\% & 88.78\% & 88.73\% & 88.17\% & 88.83\% & 83.8±0.72\% & \textbf{88.9\%} & 88.73\% \\ 
 & \textbf{Inc. ResNet V2} & 90.6±0.72\% & 92.81\% & 91.55\% & \textbf{92.89\%} & 92.57\% & 84.02±0.66\% & 91.14\% & 90.84\% \\ \midrule

\multirow{5}{*}{\textbf{Malaria}} 
 & \textbf{VGG19} & 82.0±31.99\% & \textbf{96.65\%} & 96.14\% & 96.63\% & 96.52\% & 95.86±0.01\% & 96.25\% & 96.55\% \\
 & \textbf{DenseNet 201} & 97.16±0.15\% & 97.39\% & 97.33\% & 97.33\% & \textbf{97.41\%} & 97.32±0.0\% & 97.35\% & 97.39\% \\
 & \textbf{Inception V3} & 96.95±0.2\% & \textbf{97.1\%} & 96.11\% & 97.01\% & 96.79\% & 95.89±0.02\% & 96.28\% & 96.43\% \\
 & \textbf{Xception V1} & 96.87±0.22\% & \textbf{97.04\%} & 96.39\% & 96.96\% & 96.77\% & 95.86±0.02\% & 96.48\% & 96.32\% \\ 
 & \textbf{Inc. ResNet V2} & 97.11±0.22\% & \textbf{97.32\%} & 97.21\% & 97.23\% & 97.29\% & 96.22±0.43\% & 97.31\% & 97.21\% \\ \midrule

\multirow{5}{*}{\textbf{\begin{tabular}[c]{@{}c@{}}Oxford\\ flowers\\ 102\end{tabular}}} 
 & \textbf{VGG19} & 16.03±15.03\% & \textbf{43.44\%} & 42.97\% & 38.48\% & 40.95\% & 39.94±0.39\% & 40.78\% & 39.64\% \\

 & \textbf{DenseNet 201} & 85.04±0.19\% & 67.51\% & 79.88\% & 90.59\% & 83.29\% & \textbf{92.37±2.48\%} & 84.87\% & 86.41\% \\
  & \textbf{Inception V3} & 68.6±0.73\% & 62.21\% & 71.2\% & 83.91\% & 73.21\% & \textbf{86.14±1.83\%} & 74.68\% & 74.62\% \\
 & \textbf{Xception V1} & 77.82±0.54\% & 68.18\% & 77.66\% & 88.87\% & 80.17\% & \textbf{91.88±0.48\%} & 81.3\% & 81.67\% \\ 

 & \textbf{Inc. ResNet V2} & 77.94±0.94\% & 74.54\% & 80.76\% & 86.71\% & 81.91\% & \textbf{88.83±0.26\%} & 83.21\% & 82.9\% \\ \midrule

\multirow{5}{*}{\textbf{\begin{tabular}[c]{@{}c@{}}Plant\\ leaves\end{tabular}}} 
 & \textbf{VGG19} & 93.44±1.03\% & 94.01\% & \textbf{95.15\%} & 94.82\% & 94.93\% & 94.67±0.09\% & 94.89\% & 94.73\% \\
 & \textbf{DenseNet 201} & 96.14±0.49\% & 95.28\% & 96.39\% & 96.16\% & \textbf{96.59\%} & \textbf{96.59±0.05\%} & 96.26\% & 96.18\% \\
 & \textbf{Inception V3} & 94.55±0.58\% & 94.56\% & 95.19\% & 94.58\% & \textbf{95.22\%} & 94.61±0.06\% & 94.95\% & 94.92\% \\
 & \textbf{Xception V1} & 94.63±0.38\% & 94.76\% & \textbf{95.49\%} & 94.77\% & 95.46\% & 94.9±0.09\% & 95.37\% & 95.47\% \\ 
 & \textbf{Inc. ResNet V2} & 95.81±0.42\% & 95.24\% & 95.85\% & 95.39\% & \textbf{96.02\%} & 95.6±0.14\% & 95.71\% & 95.6\% \\ \midrule

\multirow{5}{*}{\textbf{\begin{tabular}[c]{@{}c@{}}Plant\\ village\end{tabular}}} 
 & \textbf{VGG19} & 99.18±0.12\% & \textbf{99.46\%} & 99.43\% & 99.44\% & 99.43\% & 97.05±0.32\% & 99.43\% & 99.43\% \\
 & \textbf{DenseNet 201} & 99.58±0.08\% & \textbf{99.79\%} & 99.76\% & 99.75\% & 99.77\% & 97.28±0.2\% & 99.76\% & 99.76\% \\
 & \textbf{Inception V3} & 99.32±0.07\% & 99.42\% & 99.39\% & 99.35\% & \textbf{99.43\%} & 96.89±0.15\% & 99.42\% & 99.4\% \\
 & \textbf{Xception V1} & 99.41±0.08\% & \textbf{99.64\%} & 99.62\% & 99.6\% & 99.63\% & 97.12±0.29\% & 99.62\% & 99.62\% \\ 
 & \textbf{Inc. ResNet V2} & 99.44±0.13\% & 99.57\% & \textbf{99.59\%} & 99.4\% & 99.56\% & 97.1±0.23\% & 99.58\% & 99.56\% \\ \bottomrule
 
\end{tabular}%
}
\caption{Results obtained from the combination of the probabilities of each CNN model and the probabilities of the best classification algorithm trained with statistical features. Both vectors are combined using again classical supervised classification algorithms and the average between probability vectors. The classification approach with the best result for each convolutional architecture and dataset is highlighted in bold. All values indicate the macro average weighted precision in the test partition.}
\label{tab:results_comparison}
\end{table*}

\subsection{Fusion of CNN and statistical indicators-based classification}

As it was described during the description of the results obtained with the base CNNs in Section~\ref{sec:cnn_results}, there are important differences in the performance for all the architectures and domains tested, evidencing that there is room for improvement. The goal of this section is to demonstrate that the performance of CNNs for image classification can be enhanced by fusing its output with a classification model trained with statistical indicators manually extracted. 

For this purpose, we evaluate each CNN model together with the classification algorithm which showed the best performance in each domain as shown in previous Section~\ref{sec:statistical_results}. The output vectors of these two sources are combined using a new supervised classifier. The results are displayed in Table~\ref{tab:results_comparison}. In this case, since both vectors represent the same type of information (the probability of each input image to be classified as one of the possible labels), we also consider the average between both probability vectors.

In all cases, a combination of CNN probabilities and statistical information exceeds the performance of the base CNN. A simple average between the output vector of the CNN and a classifier trained on statistical features is, despite its apparent simplicity, a competitive method that even shows the best results for most of the CNN architectures in three domains: Cassava, Citrus leaves and Malaria. This probability score fusion has been providing excellent results in many classifier combination problems~\cite{Fierrez2018-1} for long. In the rest of datasets, the combination between both vectors has to be conducted using a supervised classifier. Although each domain produces varied results and there is not a clear winner approach, values obtained with Random Forest are stable across datasets and CNN architectures. 

The fusion approach produces, however, considerable differences among domains and CNN architectures. For instance, in the case of Citrus leaves, only the average helps to improve the results of the base CNN, while all the classifiers tested fail at improving the CNN performance, showing lower results. The same effect occurs in the Cars 196 dataset, but in this case Random Forest is the only approach capable of improving the performance for all CNNs except VGG19 (which does not infer knowledge in this dataset). In contrast, in domains such as Deep weeds or Cats vs dogs, several classifiers allow to exceed the base performance. All this indicates that, in order to to achieve the best possible ensemble, it is required to evaluate different methods. But, if the maximum efficiency is pursued, the average or Random Forest classifier will provide high precision rates.

On a more general level, although Random Forest shows high performance in a high number of combinations, Support Vector Machine leads to higher values in Deep weeds, Malaria and Plant village. This coincides with the three datasets with a larger number of samples. Thus, due to the high computational demand required to train SVMs, we suggest to use these models only in cases where exists a high number of examples, while Random Forest will perform better and will require less computational resources in other situations.

In case of the CNN architectures, all except VGG are improved similarly in all datasets. In case of VGG, as it was shown in Section~\ref{sec:cnn_results}, it is not able to distinguish between labels in Caltech birds 2011 and Cars 196. In these cases, the increment of the precision in the ensemble approach is due to the knowledge provided by the statistical features based classifier.

In comparison to the base CNN performance, in those datasets where the CNNs shows lower results, such as Caltech birds 2011, Cars 196, Cassava or Oxford flowers 102, the proposed approach clearly provides an improvement. For instance, in Caltech birds 2011, results improve by 6\% when using DenseNet and 7\% with Inception V3, while in Inception ResNet V2 and Xception V1 results increase by 2\%, in all cases using the Random Forest classifier. In Oxford flowers, the difference is even bigger. The results obtained in this dataset with Inception ResNet V2 increase 11\% and, in the case of Inception V3, the improvement reaches 18\%.

All these results have shown that the inclusion of statistical features can help image classification approaches based on deep architectures. By extracting different estimators directly calculated from the pixels of every image, it is possible to extract useful additional information that leads to varied improvements depending on the domain. It has also to be mentioned that the huge improvement observed in certain experiments, as it is the case of Oxford flowers102, where the precision is increased up to 18\%, can be affected by the treatment of the output probabilities of the CNN architecture with a standard classifier, arising patterns that can improve the precision in specific domains. The computation of the statistical data and the training of classical classifiers provides a useful approach with a minimal computation load in comparison to the training of a CNN.

\subsection{Ablation study}

    \begin{figure*}
        \centering
        \includegraphics[width=\textwidth]{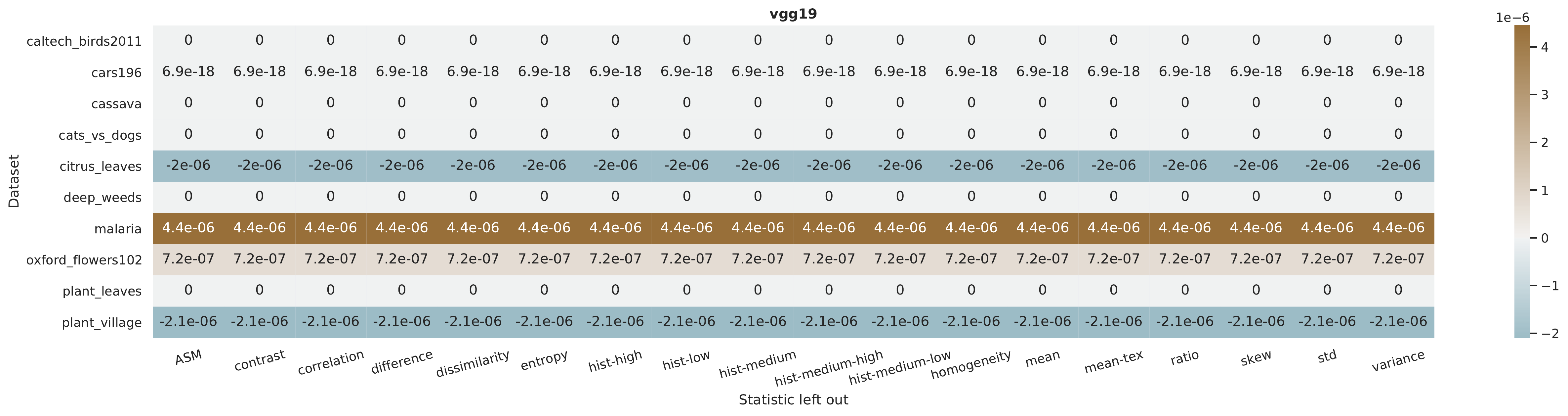}
        \caption{Results of the ablation study for the VGG19 architecture.}
        \label{fig:ablation_vgg19}
    \end{figure*}

     \begin{figure*}
        \centering
        \includegraphics[width=\textwidth]{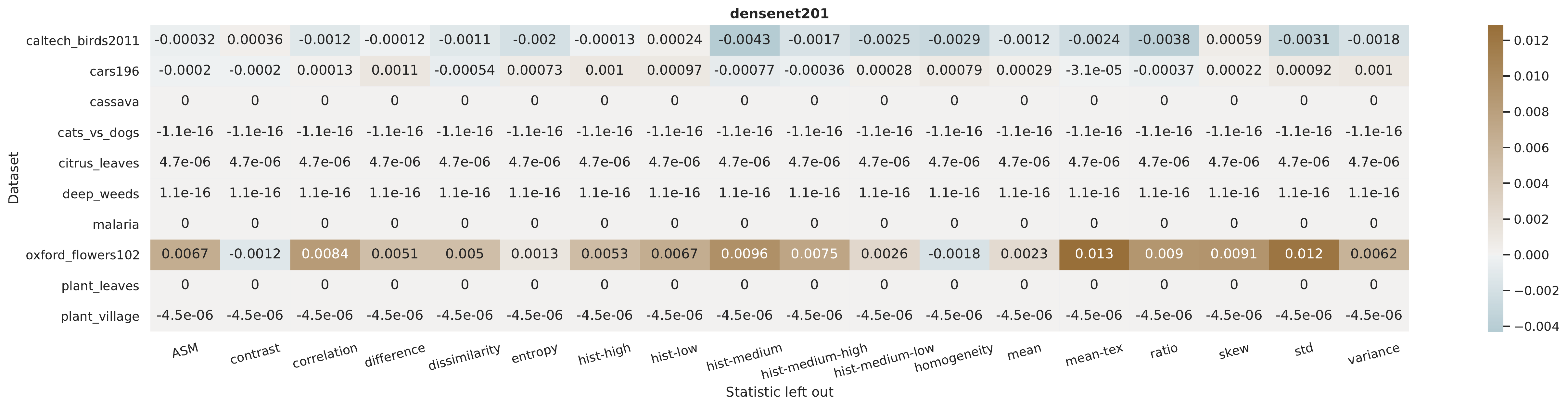}
        \caption{Results of the ablation study for the DenseNet 201 architecture.}
        \label{fig:ablation_densenet201}
     \end{figure*}

    \begin{figure*}
        \centering
        \includegraphics[width=\textwidth]{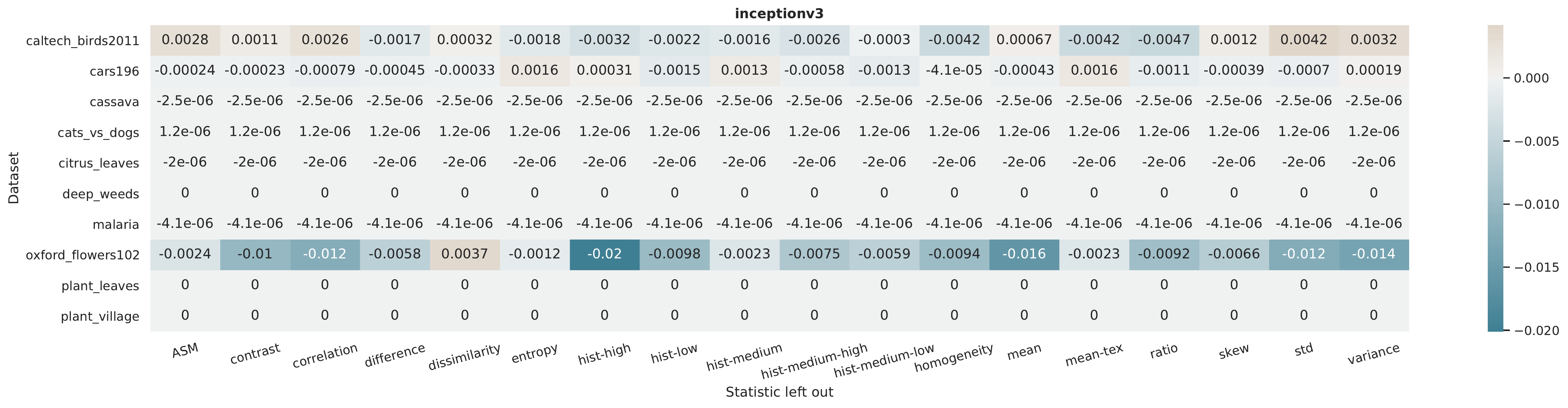}
        \caption{Results of the ablation study for the InceptionV3 architecture.}
        \label{fig:ablation_inceptionv3}
    \end{figure*}

    \begin{figure*}
        \centering
        \includegraphics[width=\textwidth]{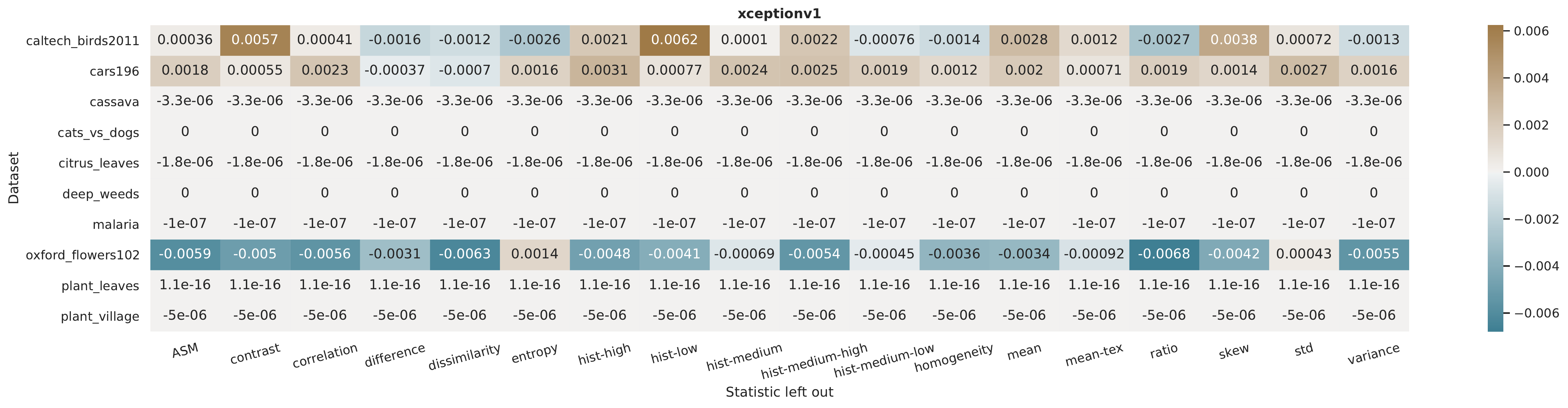}
        \caption{Results of the ablation study for the Xception V1 architecture.}
        \label{fig:ablation_xceptionv1}
    \end{figure*}

    \begin{figure*}
        \centering
        \includegraphics[width=\textwidth]{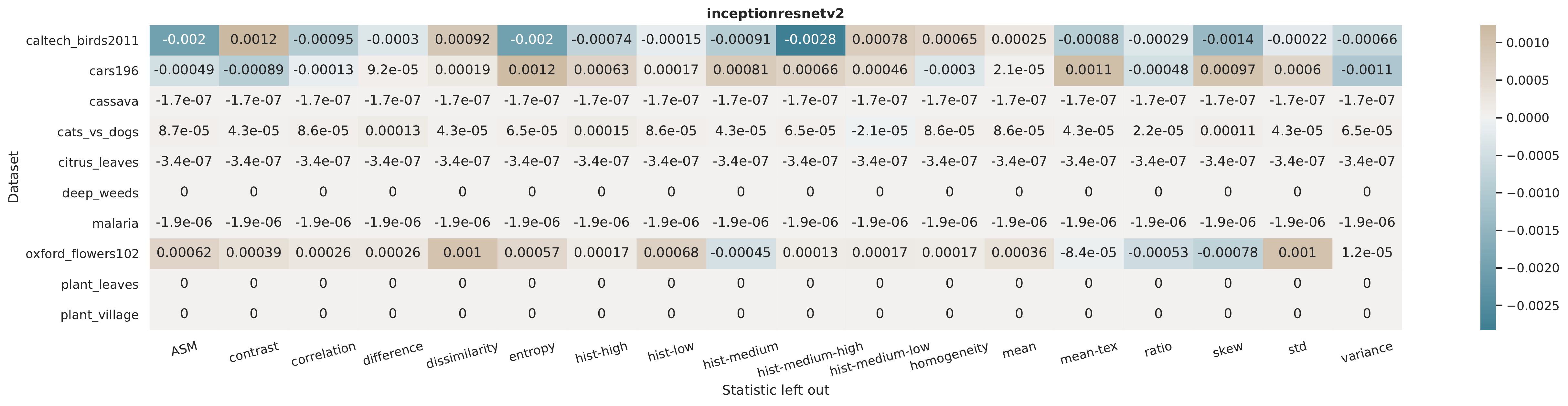}
        \caption{Results of the ablation study for the Inception ResNet V2 architecture.}
        \label{fig:ablation_inceptionresnetv2}
    \end{figure*}

In the previous section, it has been demonstrated that the addition of statistical information leads to an improvement in the classification performance in all datasets and architectures. We now assess the individual performance of each statistical indicator considered, in order quantify its contribution in the final classification and to check that all of them contribute in the same direction with no deterioration of the performance. For that purpose, this section describes the results after an ablation study addressed for each architecture and dataset. 

Figs.~\ref{fig:ablation_densenet201}, \ref{fig:ablation_inceptionresnetv2}, \ref{fig:ablation_inceptionv3}, \ref{fig:ablation_xceptionv1} and \ref{fig:ablation_vgg19} show the results of the ablation study for each CNN architecture. The goal of these figures is to describe the effect of building the fusion approach without including one statistical feature at a time. Each cell shows the difference between the ensemble approach where one specific statistical feature left out and the ensemble with all statistical features. All results were calculated after 10 different executions for both experiments. Negative values indicate a reduction of the macro average weighted precision as compared to the ensemble with all features. 

In all cases, the algorithm for the ensemble approach is selected according to the results shown in Table~\ref{tab:results_comparison}. For the sake of clarity, statistical features have been grouped by type of feature for all channels, meaning that each statistical indicator shown in the figure refers to an ensemble approach without that feature for the three channels (i.e. dissimilarity relates to three features depending on the green, red and blue channels). 

The results show high similarities across CNN architectures in the five figures. For instance, in the results for DenseNet 201 (Figure~\ref{fig:ablation_densenet201}) for the first two datasets, Caltech birds 2011 and Cars 196, it can be seen small variations between features. While the elimination of specific features slightly increase the performance, in other cases it decreases. However, the differences shown in these cells (values lower than 0,4\%) can be attributed to the use of the Random Forest classifier, the model which showed better results for all architectures except VGG19 in these two domains. 

The differences shown in the ablation study fall under the standard deviation, which are of 0.33\%, 0.42\%, 0.62\% and 0.63\% for DenseNet 201, Inception ResNet V2, Inception V3 and Xception V1 respectively (see Table~\ref{tab:results_comparison}). In the case of the VGG19 in these two datasets, there are no differences with the result obtained including all features, a fact that can be appreciated for all datasets when using VGG19. This evidences that the different statistical features extracted present strong complementarities. Nevertheless, due to the low computational resources required to extract each of them, we do not encourage to use a subset of them as they can improve the performance in each specific scenarios. For instance, in Oxford flowers 102 with Inception V3 (see Figure~\ref{fig:ablation_inceptionv3}), the elimination of hist-high causes a 2\% decrease in the performance, while contrast and correlation leads to a 1\% decrease, evidencing the importance of these two features.

\subsection{Time performance analysis}

An analysis based on the time performance of the proposed approach is discussed in this section. Table~\ref{tab:training_times}, provided in Appendix A, shows all the training times in minutes for the different parts of the method proposed, for all datasets, CNN architectures and ensemble algorithms.

As expected, the CNN is the most time-consuming element of the approach, and it is highly dependent on the number of training examples but also in the general complexity of the domain. The feature extraction process, in contrast, is very efficient and, for most cases, last less than one minute. In one specific case, for the Plant leaves dataset, due to the characteristics of this dataset, it takes 15 minutes. However, it is still a quarter of the training time of the CNN. For the statistical features-based classification, the time required is also minimal in comparison to the CNN. This is caused by the low size of the feature space, consisting of 54 different characteristics. In this case, the number of training examples will increase the training time accordingly. 

The final ensemble approach is also very efficient. However, in this case, the use of Support Vector Machines causes huge training times in several datasets (Caltech birds 2011, Cars 196 or Plant village). This is caused due to the number of labels but also due to the number of training examples. If maximum efficiency is pursued, we recommend to avoid this type of classifier, as those based on decision trees or Linear Discriminant Analysis will provide high precision rates more efficiently. In any case, the minimum training time is obtained using the average between both probability vectors which, as previously showed, achieves high values in several datasets and architectures.


\section{Conclusions and future works}\label{section:conclusions}

On this article a novel approach to improve CNNs has been proposed. This approach consists in combining traditional manually extracted features with the automatic feature extraction of Convolutional Neural Networks. This ensemble model contains a CNN module that labels examples with a probability vector, a statistical-features based classifier that labels examples with either a probability vector or a one-hot encoded vector and, finally, both vectors are combined and re-classified through a fusion approach to get the final labelling of the image. This method has been tested with several CNN algorithms on heterogeneous datasets to demonstrate the general capabilities of the method. 

The experimentation show promising results, as the proposed additions are never detrimental independently of the domain or CNN network applied. Although the improvement is subject to the performance of the CNN, which implies that in those domains where the CNN reaches high accuracy rates there will be little space for improvement, in all domains tested the approach is able to improve the results. Besides, training the feature classifier and fusion classifier modules is much cheaper than training upscaled versions of the studied networks. As we show in the Appendix, the time required to train both classifiers is minimal in comparison to the training of the CNN. It is also worth noting that some feature classifiers (and fusion classifiers) have better explainability than the CNN~\cite{paco2020xai,2021_WACV_XAI_Alfonso}, although these explainable models are not always the best option.

This proposal could be improved in several ways in future work. For example, manually extracted features could be further improved by using new approaches to feature extraction including more advanced texture detection techniques, or hand-crafted feature extraction techniques specifically developed for the domain at hand.

Other venues for future research may include using boosting techniques in the fusion classifier such as extreme gradient boosting, among others. The fusion classifier can be also improved by first designing it using large-scale development data representing the general probability behaviors in wide domains, and then fine-tuning it to specific problems or application areas in a kind of adapted fusion scheme~\cite{Fierrez2018-2,FierrezAguilar2005}.

An interesting research line could appear from the study and comparison of machine learning extracted features and manually extracted features as indicated by previous research~\cite{Huertas_Tato_Martin_Camacho_2020}. Our results show that the predictions of CNNs and other algorithms are fundamentally different, and that their combination is greatly beneficial to performance. These differences could be key to understand and explain why CNNs fail to classify some examples on specific domains. This example-based behavior that makes CNNs to fail while other simpler classifiers can work much better can be exploited with context-based fusion classifiers~\cite{laura2015context}, e.g., switching between them depending on the input images and context~\cite{Alonso2010}, or combining them with example-dependent adaptive fusion schemes~\cite{Fierrez2018-2}. Finally, we also plan to adapt the proposed ensemble approach to reduce undesired biases \cite{pena20bias,serna21bias} in pre-trained networks.

\section{Acknowledgements*}

This work has been supported partially supported by following grants and funding agencies: Spanish Ministry of Science and Innovation under TIN2017-85727-C4-3-P (DeepBio) and RTI2018-101248-B-I00 (BIBECA) grants, by Comunidad Aut\'{o}noma de Madrid under S2018/TCS-4566 grant (CYNAMON), and by BBVA FOUNDATION GRANTS FOR SCIENTIFIC RESEARCH TEAMS SARS-CoV-2 and COVID-19 under the grant: "\textit{CIVIC: Intelligent characterisation of the veracity of the information related to COVID-19}". Finally, the work has been supported by the Comunidad Aut\'{o}noma de Madrid under: "Convenio Plurianual with the Universidad Politécnica de Madrid in the actuation line of \textit{Programa de Excelencia para el Profesorado Universitario}".




\bibliographystyle{unsrt}  
\bibliography{bibtex, bibtex-new}   

\begin{thebibliography}{10}

\bibitem{lecun1995convolutional}
Yann LeCun, Yoshua Bengio, et~al.
\newblock Convolutional networks for images, speech, and time series.
\newblock {\em The handbook of brain theory and neural networks},
  3361(10):1995, 1995.

\bibitem{liu2017survey}
Weibo Liu, Zidong Wang, Xiaohui Liu, Nianyin Zeng, Yurong Liu, and Fuad~E
  Alsaadi.
\newblock A survey of deep neural network architectures and their applications.
\newblock {\em Neurocomputing}, 234:11--26, 2017.

\bibitem{Brown2020May}
Tom~B. Brown, Benjamin Mann, Nick Ryder, Melanie Subbiah, Jared Kaplan,
  Prafulla Dhariwal, Arvind Neelakantan, Pranav Shyam, Girish Sastry, Amanda
  Askell, Sandhini Agarwal, Ariel Herbert-Voss, Gretchen Krueger, Tom Henighan,
  Rewon Child, Aditya Ramesh, Daniel~M. Ziegler, Jeffrey Wu, Clemens Winter,
  Christopher Hesse, Mark Chen, Eric Sigler, Mateusz Litwin, Scott Gray,
  Benjamin Chess, Jack Clark, Christopher Berner, Sam McCandlish, Alec Radford,
  Ilya Sutskever, and Dario Amodei.
\newblock {Language Models are Few-Shot Learners}.
\newblock {\em arXiv}, May 2020.

\bibitem{simonyan2014very}
Karen Simonyan and Andrew Zisserman.
\newblock {Very Deep Convolutional Networks for Large-Scale Image Recognition}.
\newblock {\em arXiv}, Sep 2014.

\bibitem{Szegedy2015Dec}
C.~{Szegedy}, V.~{Vanhoucke}, S.~{Ioffe}, J.~{Shlens}, and Z.~{Wojna}.
\newblock Rethinking the inception architecture for computer vision.
\newblock In {\em 2016 IEEE Conference on Computer Vision and Pattern
  Recognition (CVPR)}, pages 2818--2826, 2016.

\bibitem{Real2018Feb}
Esteban Real, Alok Aggarwal, Yanping Huang, and Quoc~V. Le.
\newblock {Regularized Evolution for Image Classifier Architecture Search}.
\newblock {\em arXiv}, Feb 2018.

\bibitem{Schwartz2019Jul}
Roy Schwartz, Jesse Dodge, Noah~A. Smith, and Oren Etzioni.
\newblock {Green AI}.
\newblock {\em arXiv}, Jul 2019.

\bibitem{strubell_energy_2020}
Emma Strubell, Ananya Ganesh, and Andrew McCallum.
\newblock Energy and {Policy} {Considerations} for {Modern} {Deep} {Learning}
  {Research}.
\newblock {\em Proceedings of the AAAI Conference on Artificial Intelligence},
  34(09):13693--13696, April 2020.

\bibitem{Steinkraus2005Aug}
D.~{Steinkraus}, I.~{Buck}, and P.~Y. {Simard}.
\newblock Using gpus for machine learning algorithms.
\newblock In {\em Eighth International Conference on Document Analysis and
  Recognition (ICDAR'05)}, pages 1115--1120 Vol. 2, 2005.

\bibitem{xn--Bucilu-85b2006Aug}
Cristian Bucilua, Rich Caruana, and Alexandru Niculescu-Mizil.
\newblock {\em {Model compression}}.
\newblock Association for Computing Machinery, New York, NY, USA, Aug 2006.

\bibitem{Cheng2017Oct}
Yu~Cheng, Duo Wang, Pan Zhou, and Tao Zhang.
\newblock {A Survey of Model Compression and Acceleration for Deep Neural
  Networks}.
\newblock {\em arXiv}, Oct 2017.

\bibitem{martin2020statistically}
Alejandro Martin, Ra{\'u}l Lara-Cabrera, V{\'\i}ctor~Manuel Vargas,
  Pedro~Antonio Guti{\'e}rrez, C{\'e}sar Herv{\'a}s-Mart{\'\i}nez, and David
  Camacho.
\newblock Statistically-driven coral reef metaheuristic for automatic
  hyperparameter setting and architecture design of convolutional neural
  networks.
\newblock In {\em 2020 IEEE Congress on Evolutionary Computation (CEC)}, pages
  1--8. IEEE, 2020.

\bibitem{Hinton2015Mar}
Geoffrey Hinton, Oriol Vinyals, and Jeff Dean.
\newblock {Distilling the Knowledge in a Neural Network}.
\newblock {\em arXiv}, Mar 2015.

\bibitem{He2017}
Yihui He, Xiangyu Zhang, and Jian Sun.
\newblock {Channel Pruning for Accelerating Very Deep Neural Networks}.
\newblock In {\em 2017 IEEE International Conference on Computer Vision
  (ICCV)}, pages 1389--1397, 2017.
\newblock [Online; accessed 20. Oct. 2020].

\bibitem{Luo2017}
Jian-Hao Luo, Jianxin Wu, and Weiyao Lin.
\newblock Thinet: A filter level pruning method for deep neural network
  compression.
\newblock In {\em Proceedings of the IEEE international conference on computer
  vision}, pages 5058--5066, 2017.

\bibitem{martin2020optimising}
Alejandro Mart{\'\i}n, V{\'\i}ctor~Manuel Vargas, Pedro~Antonio Guti{\'e}rrez,
  David Camacho, and C{\'e}sar Herv{\'a}s-Mart{\'\i}nez.
\newblock Optimising convolutional neural networks using a hybrid
  statistically-driven coral reef optimisation algorithm.
\newblock {\em Applied Soft Computing}, 90:106144, 2020.

\bibitem{Cohen2016Feb}
Taco~S. Cohen and Max Welling.
\newblock {Group Equivariant Convolutional Networks}.
\newblock {\em arXiv}, Feb 2016.

\bibitem{Iandola2016Feb}
Forrest~N. Iandola, Song Han, Matthew~W. Moskewicz, Khalid Ashraf, William~J.
  Dally, and Kurt Keutzer.
\newblock {SqueezeNet: AlexNet-level accuracy with 50x fewer parameters and
  {$<$}0.5MB model size}.
\newblock {\em arXiv}, Feb 2016.

\bibitem{martin2018evodeep}
Alejandro Mart{\'\i}n, Ra{\'u}l Lara-Cabrera, F{\'e}lix Fuentes-Hurtado, Valery
  Naranjo, and David Camacho.
\newblock Evodeep: a new evolutionary approach for automatic deep neural
  networks parametrisation.
\newblock {\em Journal of Parallel and Distributed Computing}, 117:180--191,
  2018.

\bibitem{Tan2019May}
Mingxing Tan and Quoc~V. Le.
\newblock {EfficientNet: Rethinking Model Scaling for Convolutional Neural
  Networks}.
\newblock {\em arXiv}, May 2019.

\bibitem{Zoph2017Jul}
Barret Zoph, Vijay Vasudevan, Jonathon Shlens, and Quoc~V. Le.
\newblock {Learning Transferable Architectures for Scalable Image Recognition}.
\newblock {\em arXiv}, Jul 2017.

\bibitem{Huang2018Nov}
Yanping Huang, Youlong Cheng, Ankur Bapna, Orhan Firat, Mia~Xu Chen, Dehao
  Chen, HyoukJoong Lee, Jiquan Ngiam, Quoc~V. Le, Yonghui Wu, and Zhifeng Chen.
\newblock {GPipe: Efficient Training of Giant Neural Networks using Pipeline
  Parallelism}.
\newblock {\em arXiv}, Nov 2018.

\bibitem{Zhang2018}
Xiangyu Zhang, Xinyu Zhou, Mengxiao Lin, and Jian Sun.
\newblock {ShuffleNet: An Extremely Efficient Convolutional Neural Network for
  Mobile Devices}, 2018.

\bibitem{Howard2017Apr}
Andrew~G. Howard, Menglong Zhu, Bo~Chen, Dmitry Kalenichenko, Weijun Wang,
  Tobias Weyand, Marco Andreetto, and Hartwig Adam.
\newblock {MobileNets: Efficient Convolutional Neural Networks for Mobile
  Vision Applications}.
\newblock {\em arXiv}, Apr 2017.

\bibitem{Fierrez2018-1}
Julian Fierrez, Aythami Morales, Ruben Vera-Rodriguez, and David Camacho.
\newblock Multiple classifiers in biometrics. part 1: Fundamentals and review.
\newblock {\em Information Fusion}, 44:57--64, November 2018.

\bibitem{Fierrez2018-2}
Julian Fierrez, Aythami Morales, Ruben Vera-Rodriguez, and David Camacho.
\newblock Multiple classifiers in biometrics. part 2: Trends and challenges.
\newblock {\em Information Fusion}, 44:103--112, November 2018.

\bibitem{Roy2020}
Swalpa~Kumar Roy, Shiv~Ram Dubey, Bhabatosh Chanda, Bidyut~B. Chaudhuri, and
  Dipak~Kumar Ghosh.
\newblock {TexFusionNet: An Ensemble of Deep CNN Feature for Texture
  Classification}.
\newblock In {\em {Proceedings of 3rd International Conference on Computer
  Vision and Image Processing}}, pages 271--283, Singapore, Sep 2019. Springer.

\bibitem{krizhevsky2017imagenet}
Alex Krizhevsky, Ilya Sutskever, and Geoffrey~E Hinton.
\newblock Imagenet classification with deep convolutional neural networks.
\newblock {\em Communications of the ACM}, 60(6):84--90, 2017.

\bibitem{Amin-Naji2019Nov}
Mostafa Amin-Naji, Ali Aghagolzadeh, and Mehdi Ezoji.
\newblock {Ensemble of CNN for multi-focus image fusion}.
\newblock {\em Information Fusion}, 51:201--214, Nov 2019.

\bibitem{Deng2019Nov}
Pan Deng, Haipeng Chen, Mengyao Huang, Xiaowen Ruan, and Liang Xu.
\newblock {An ensemble CNN method for biomedical entity normalization}.
\newblock {\em ACL Anthology}, pages 143--149, Nov 2019.

\bibitem{Huang2006Jun}
{Fu Jie Huang} and Y.~{LeCun}.
\newblock Large-scale learning with svm and convolutional for generic object
  categorization.
\newblock In {\em 2006 IEEE Computer Society Conference on Computer Vision and
  Pattern Recognition (CVPR'06)}, volume~1, pages 284--291, 2006.

\bibitem{Niu2012Apr}
Xiao-Xiao Niu and Ching~Y. Suen.
\newblock {A novel hybrid CNN{\textendash}SVM classifier for recognizing
  handwritten digits}.
\newblock {\em Pattern Recognit.}, 45(4):1318--1325, Apr 2012.

\bibitem{liz2021ensembles}
Helena Liz, Manuel S{\'a}nchez-Monta{\~n}{\'e}s, Alfredo Tagarro, Sara
  Dom{\'\i}nguez-Rodr{\'\i}guez, Ron Dagan, and David Camacho.
\newblock Ensembles of convolutional neural network models for pediatric
  pneumonia diagnosis.
\newblock {\em Future Generation Computer Systems}, 122:220--233, 2021.

\bibitem{martin2019android}
Alejandro Mart{\'\i}n, Raul Lara-Cabrera, and David Camacho.
\newblock Android malware detection through hybrid features fusion and ensemble
  classifiers: The andropytool framework and the omnidroid dataset.
\newblock {\em Information Fusion}, 52:128--142, 2019.

\bibitem{Nixon2012Oct}
Mark Nixon and Alberto~S. Aguado.
\newblock {\em {Feature Extraction {\&} Image Processing for Computer Vision,
  Third Edition}}.
\newblock Academic Press, Inc.6277 Sea Harbor Drive Orlando, FLUnited States,
  Oct 2012.

\bibitem{storcheus2015survey}
Dmitry Storcheus, Afshin Rostamizadeh, and Sanjiv Kumar.
\newblock A survey of modern questions and challenges in feature extraction.
\newblock In {\em Feature Extraction: Modern Questions and Challenges}, pages
  1--18, 2015.

\bibitem{Khalid2014Aug}
Samina Khalid, Tehmina Khalil, and Shamila Nasreen.
\newblock {A survey of feature selection and feature extraction techniques in
  machine learning}.
\newblock {\em 2014 Science and Information Conference}, pages 372--378, Aug
  2014.

\bibitem{haralick_textural_1973}
R.~M. Haralick, I.~Dinstein, and K.~Shanmugam.
\newblock Textural {Features} for {Image} {Classification}.
\newblock {\em IEEE Transactions on Systems, Man and Cybernetics},
  SMC-3(6):610--621, 1973.

\bibitem{Ojala2002Aug}
T.~Ojala, M.~Pietikainen, and T.~Maenpaa.
\newblock {Multiresolution gray-scale and rotation invariant texture
  classification with local binary patterns}.
\newblock {\em IEEE Trans. Pattern Anal. Mach. Intell.}, 24(7):971--987, Aug
  2002.

\bibitem{cortes1995support}
Corinna Cortes and Vladimir Vapnik.
\newblock Support-vector networks.
\newblock {\em Machine learning}, 20(3):273--297, 1995.

\bibitem{Kim2002Nov}
Kwang~In Kim, Keechul Jung, Se~Hyun Park, and Hang~Joon Kim.
\newblock {Support vector machines for texture classification}.
\newblock {\em IEEE Trans. Pattern Anal. Mach. Intell.}, 24(11):1542--1550, Nov
  2002.

\bibitem{Sosa2018}
Ester Gonzalez-Sosa, Julian Fierrez, Ruben Vera-Rodriguez, and Fernando
  Alonso-Fernandez.
\newblock Facial soft biometrics for recognition in the wild: Recent works,
  annotation and cots evaluation.
\newblock {\em IEEE Trans. on Information Forensics and Security},
  13(8):2001--2014, August 2018.

\bibitem{Sun2020Oct}
{Zehang Sun}, G.~{Bebis}, and R.~{Miller}.
\newblock On-road vehicle detection using gabor filters and support vector
  machines.
\newblock In {\em 2002 14th International Conference on Digital Signal
  Processing Proceedings. DSP 2002 (Cat. No.02TH8628)}, volume~2, pages
  1019--1022 vol.2, 2002.

\bibitem{Faundez2021}
Marcos Faundez-Zanuy, Julian Fierrez, Miguel~A. Ferrer, Moises Diaz, Ruben
  Tolosana, and RÃ©jean Plamondon.
\newblock Handwriting biometrics: Applications and future trends in e-security
  and e-health.
\newblock {\em Cognitive Computation}, August 2020.

\bibitem{ramteke2012automatic}
RJ~Ramteke and Khachane~Y Monali.
\newblock Automatic medical image classification and abnormality detection
  using k-nearest neighbour.
\newblock {\em International Journal of Advanced Computer Research}, 2(4):190,
  2012.

\bibitem{Munisami2015Jan}
Trishen Munisami, Mahess Ramsurn, Somveer Kishnah, and Sameerchand Pudaruth.
\newblock {Plant Leaf Recognition Using Shape Features and Colour Histogram
  with K-nearest Neighbour Classifiers}.
\newblock {\em Procedia Comput. Sci.}, 58:740--747, Jan 2015.

\bibitem{Rajini2020}
N.~H. {Rajini} and R.~{Bhavani}.
\newblock Classification of mri brain images using k-nearest neighbor and
  artificial neural network.
\newblock In {\em 2011 International Conference on Recent Trends in Information
  Technology (ICRTIT)}, pages 563--568, 2011.

\bibitem{breiman_random_2001}
Leo Breiman.
\newblock Random {Forests}.
\newblock {\em Machine Learning}, 45(1):5--32, October 2001.

\bibitem{Du2015Jul}
Peijun Du, Alim Samat, Bj{\ifmmode\ddot{o}\else\"{o}\fi}rn Waske, Sicong Liu,
  and Zhenhong Li.
\newblock {Random Forest and Rotation Forest for fully polarized SAR image
  classification using polarimetric and spatial features}.
\newblock {\em ISPRS J. Photogramm. Remote Sens.}, 105:38--53, Jul 2015.

\bibitem{Mishra2017Mar}
Sonali Mishra, Banshidhar Majhi, Pankaj~Kumar Sa, and Lokesh Sharma.
\newblock {Gray level co-occurrence matrix and random forest based acute
  lymphoblastic leukemia detection}.
\newblock {\em Biomed. Signal Process. Control}, 33:272--280, Mar 2017.

\bibitem{Gray2013Jan}
Katherine~R. Gray, Paul Aljabar, Rolf~A. Heckemann, Alexander Hammers, and
  Daniel Rueckert.
\newblock {Random forest-based similarity measures for multi-modal
  classification of Alzheimer's disease}.
\newblock {\em Neuroimage}, 65:167--175, Jan 2013.

\bibitem{ding2021ap}
Yifeng Ding, Zhanyu Ma, Shaoguo Wen, Jiyang Xie, Dongliang Chang, Zhongwei Si,
  Ming Wu, and Haibin Ling.
\newblock Ap-cnn: weakly supervised attention pyramid convolutional neural
  network for fine-grained visual classification.
\newblock {\em IEEE Transactions on Image Processing}, 30:2826--2836, 2021.

\bibitem{chang2020devil}
Dongliang Chang, Yifeng Ding, Jiyang Xie, Ayan~Kumar Bhunia, Xiaoxu Li, Zhanyu
  Ma, Ming Wu, Jun Guo, and Yi-Zhe Song.
\newblock The devil is in the channels: Mutual-channel loss for fine-grained
  image classification.
\newblock {\em IEEE Transactions on Image Processing}, 29:4683--4695, 2020.

\bibitem{peng2019fb}
Yingqiong Peng, Muxin Liao, Yuxia Song, Zhichao Liu, Huojiao He, Hong Deng, and
  Yinglong Wang.
\newblock Fb-cnn: Feature fusion-based bilinear cnn for classification of fruit
  fly image.
\newblock {\em IEEE Access}, 8:3987--3995, 2019.

\bibitem{fierrez06phd}
Julian Fierrez.
\newblock {\em Adapted Fusion Schemes for Multimodal Biometric Authentication}.
\newblock PhD thesis, Universidad Politecnica de Madrid, 5 2006.

\bibitem{heinle_automatic_2010}
A.~Heinle, A.~Macke, and A.~Srivastav.
\newblock Automatic cloud classification of whole sky images.
\newblock {\em Atmospheric Measurement Techniques}, 3(3):557--567, 2010.

\bibitem{Pu_Sun_Ma_Cheng_2015}
Hongbin Pu, Da-Wen Sun, Ji~Ma, and Jun-Hu Cheng.
\newblock Classification of fresh and frozen-thawed pork muscles using visible
  and near infrared hyperspectral imaging and textural analysis.
\newblock {\em Meat Science}, 99:81–88, 2015.

\bibitem{Tuominen_Pekkarinen_2005}
Sakari Tuominen and Anssi Pekkarinen.
\newblock Performance of different spectral and textural aerial photograph
  features in multi-source forest inventory.
\newblock {\em Remote sensing of Environment}, 94(2):256–268, 2005.

\bibitem{Venkataraman_Mangayarkarasi_2016}
D~Venkataraman and N~Mangayarkarasi.
\newblock Computer vision based feature extraction of leaves for identification
  of medicinal values of plants.
\newblock In {\em 2016 IEEE International Conference on Computational
  Intelligence and Computing Research (ICCIC)}, page 1–5. IEEE, 2016.

\bibitem{jensen1979spectral}
John~R Jensen.
\newblock Spectral and textural features to classify elusive land cover at the
  urban fringe.
\newblock {\em The Professional Geographer}, 31(4):400--409, 1979.

\bibitem{pena2014object}
Jos{\'e}~M Pe{\~n}a, Pedro~A Guti{\'e}rrez, C{\'e}sar Herv{\'a}s-Mart{\'\i}nez,
  Johan Six, Richard~E Plant, and Francisca L{\'o}pez-Granados.
\newblock Object-based image classification of summer crops with machine
  learning methods.
\newblock {\em Remote Sensing}, 6(6):5019--5041, 2014.

\bibitem{zhang2017image}
Huanxue Zhang, Qiangzi Li, Jiangui Liu, Jiali Shang, Xin Du, Heather McNairn,
  Catherine Champagne, Taifeng Dong, and Mingxu Liu.
\newblock Image classification using rapideye data: Integration of spectral and
  textual features in a random forest classifier.
\newblock {\em IEEE Journal of Selected Topics in Applied Earth Observations
  and Remote Sensing}, 10(12):5334--5349, 2017.

\bibitem{scikit-learn}
F.~Pedregosa, G.~Varoquaux, A.~Gramfort, V.~Michel, B.~Thirion, O.~Grisel,
  M.~Blondel, P.~Prettenhofer, R.~Weiss, V.~Dubourg, J.~Vanderplas, A.~Passos,
  D.~Cournapeau, M.~Brucher, M.~Perrot, and E.~Duchesnay.
\newblock Scikit-learn: Machine learning in {P}ython.
\newblock {\em Journal of Machine Learning Research}, 12:2825--2830, 2011.

\bibitem{fletcher2013practical}
Roger Fletcher.
\newblock {\em Practical methods of optimization}.
\newblock John Wiley \& Sons, 2013.

\bibitem{WelinderEtal2010}
P.~Welinder, S.~Branson, T.~Mita, C.~Wah, F.~Schroff, S.~Belongie, and
  P.~Perona.
\newblock {Caltech-UCSD Birds 200}.
\newblock Technical Report CNS-TR-2010-001, California Institute of Technology,
  2010.

\bibitem{KrauseStarkDengFei-Fei_3DRR2013}
Jonathan Krause, Michael Stark, Jia Deng, and Li~Fei-Fei.
\newblock 3d object representations for fine-grained categorization.
\newblock In {\em 4th International IEEE Workshop on 3D Representation and
  Recognition (3dRR-13)}, Sydney, Australia, 2013.

\bibitem{mwebaze2019icassava}
Ernest Mwebaze, Timnit Gebru, Andrea Frome, Solomon Nsumba, and Jeremy
  Tusubira.
\newblock icassava 2019fine-grained visual categorization challenge, 2019.

\bibitem{asirra2007elson}
Jeremy Elson, John~(JD) Douceur, Jon Howell, and Jared Saul.
\newblock Asirra: A captcha that exploits interest-aligned manual image
  categorization.
\newblock In {\em Proceedings of 14th ACM Conference on Computer and
  Communications Security (CCS)}. Association for Computing Machinery, Inc.,
  October 2007.

\bibitem{rauf2019citrus}
Hafiz~Tayyab Rauf, Basharat~Ali Saleem, M~Ikram~Ullah Lali, Muhammad~Attique
  Khan, Muhammad Sharif, and Syed Ahmad~Chan Bukhari.
\newblock A citrus fruits and leaves dataset for detection and classification
  of citrus diseases through machine learning.
\newblock {\em Data in brief}, 26:104340, 2019.

\bibitem{DeepWeeds2019}
Alex Olsen, Dmitry~A. Konovalov, Bronson Philippa, Peter Ridd, Jake~C. Wood,
  Jamie Johns, Wesley Banks, Benjamin Girgenti, Owen Kenny, James Whinney,
  Brendan Calvert, Mostafa {Rahimi Azghadi}, and Ronald~D. White.
\newblock {DeepWeeds: A Multiclass Weed Species Image Dataset for Deep
  Learning}.
\newblock {\em Scientific Reports}, 9(2058), 2 2019.

\bibitem{rajaraman2018pre}
Sivaramakrishnan Rajaraman, Sameer~K Antani, Mahdieh Poostchi, Kamolrat
  Silamut, Md~A Hossain, Richard~J Maude, Stefan Jaeger, and George~R Thoma.
\newblock Pre-trained convolutional neural networks as feature extractors
  toward improved malaria parasite detection in thin blood smear images.
\newblock {\em PeerJ}, 6:e4568, 2018.

\bibitem{Nilsback08}
M-E. Nilsback and A.~Zisserman.
\newblock Automated flower classification over a large number of classes.
\newblock In {\em Proceedings of the Indian Conference on Computer Vision,
  Graphics and Image Processing}, Dec 2008.

\bibitem{plant2019siddhart}
Siddharth~Singh Chouhan, Ajay Kaul, Uday~Pratap Singh, and Sanjeev Jain.
\newblock {A Database of Leaf Images: Practice towards Plant Conservation with
  Plant Pathology}.
\newblock {\em mendeley}, 1, Jun 2019.

\bibitem{DBLP:journals/corr/HughesS15}
David~P. Hughes and Marcel Salath{'{e} }.
\newblock An open access repository of images on plant health to enable the
  development of mobile disease diagnostics through machine learning and
  crowdsourcing.
\newblock {\em CoRR}, abs/1511.08060, 2015.

\bibitem{Huang2017Jul}
Gao Huang, Zhuang Liu, Laurens Van Der~Maaten, and Kilian~Q. Weinberger.
\newblock {Densely Connected Convolutional Networks}.
\newblock {\em 2017 IEEE Conference on Computer Vision and Pattern Recognition
  (CVPR)}, pages 2261--2269, Jul 2017.

\bibitem{He2015Dec}
Kaiming He, Xiangyu Zhang, Shaoqing Ren, and Jian Sun.
\newblock {Deep Residual Learning for Image Recognition}.
\newblock {\em arXiv}, Dec 2015.

\bibitem{Szegedy2016Feb}
Christian Szegedy, Sergey Ioffe, Vincent Vanhoucke, and Alex Alemi.
\newblock {Inception-v4, Inception-ResNet and the Impact of Residual
  Connections on Learning}.
\newblock {\em arXiv}, Feb 2016.

\bibitem{chollet2015keras}
Fran\c{c}ois Chollet et~al.
\newblock Keras.
\newblock \url{https://keras.io}, 2015.

\bibitem{tensorflow2015-whitepaper}
Mart\'{\i}n Abadi, Ashish Agarwal, Paul Barham, Eugene Brevdo, Zhifeng Chen,
  Craig Citro, Greg~S. Corrado, Andy Davis, Jeffrey Dean, Matthieu Devin,
  Sanjay Ghemawat, Ian Goodfellow, Andrew Harp, Geoffrey Irving, Michael Isard,
  Yangqing Jia, Rafal Jozefowicz, Lukasz Kaiser, Manjunath Kudlur, Josh
  Levenberg, Dan Man\'{e}, Rajat Monga, Sherry Moore, Derek Murray, Chris Olah,
  Mike Schuster, Jonathon Shlens, Benoit Steiner, Ilya Sutskever, Kunal Talwar,
  Paul Tucker, Vincent Vanhoucke, Vijay Vasudevan, Fernanda Vi\'{e}gas, Oriol
  Vinyals, Pete Warden, Martin Wattenberg, Martin Wicke, Yuan Yu, and Xiaoqiang
  Zheng.
\newblock {TensorFlow}: Large-scale machine learning on heterogeneous systems,
  2015.
\newblock Software available from tensorflow.org.

\bibitem{paco2020xai}
Alejandro {Barredo Arrieta}, Natalia Díaz-Rodríguez, Javier {Del Ser}, Adrien
  Bennetot, Siham Tabik, Alberto Barbado, Salvador Garcia, Sergio Gil-Lopez,
  Daniel Molina, Richard Benjamins, Raja Chatila, and Francisco Herrera.
\newblock Explainable artificial intelligence (xai): Concepts, taxonomies,
  opportunities and challenges toward responsible ai.
\newblock {\em Information Fusion}, 58:82 -- 115, 2020.

\bibitem{2021_WACV_XAI_Alfonso}
Alfonso Ortega, Julian Fierrez, Aythami Morales, Zilong Wang, and Tony Ribeiro.
\newblock Symbolic ai for xai: Evaluating lfit inductive programming for fair
  and explainable automatic recruitment.
\newblock In {\em IEEE/CVF Winter Conf. on Applications of Computer Vision
  Workshops (WACVw)}, January 2021.

\bibitem{FierrezAguilar2005}
J.~Fierrez-Aguilar, D.~Garcia-Romero, J.~Ortega-Garcia, and
  J.~Gonzalez-Rodriguez.
\newblock Adapted user-dependent multimodal biometric authentication exploiting
  general information.
\newblock {\em Pattern Recognition Letters}, 26(16):2628--2639, December 2005.

\bibitem{Huertas_Tato_Martin_Camacho_2020}
Javier Huertas-Tato, Alejandro Martín, and David Camacho.
\newblock {\em Cloud Type Identification Using Data Fusion and Ensemble
  Learning}, volume 12490 of {\em Lecture Notes in Computer Science}, page
  137–147.
\newblock Springer International Publishing, 2020.

\bibitem{laura2015context}
Lauro Snidaro, Jesús García, and James Llinas.
\newblock Context-based information fusion: A survey and discussion.
\newblock {\em Information Fusion}, 25:16 -- 31, 2015.

\bibitem{Alonso2010}
Fernando Alonso-Fernandez, Julian Fierrez, Daniel Ramos, and Joaquin
  Gonzalez-Rodriguez.
\newblock Quality-based conditional processing in multi-biometrics: application
  to sensor interoperability.
\newblock {\em IEEE Trans. on Systems, Man and Cybernetics Part A},
  40(6):1168--1179, 2010.

\bibitem{pena20bias}
Alejandro Pena, Ignacio Serna, Aythami Morales, and Julian Fierrez.
\newblock Bias in multimodal ai: Testbed for fair automatic recruitment.
\newblock In {\em IEEE/CVF Conf. on Computer Vision and Pattern Recognition
  Workshops (CVPRw)}, June 2020.
\newblock also at ICML 2020 Workshop on Human-in-the-Loop Learning.

\bibitem{serna21bias}
Ignacio Serna, Alejandro Pena, Aythami Morales, and Julian Fierrez.
\newblock Insidebias: Measuring bias in deep networks and application to face
  gender biometrics.
\newblock In {\em IAPR Intl. Conf. on Pattern Recognition (ICPR)}, January
  2021.

\end{thebibliography}
\clearpage
\onecolumn
\appendix

\section{Time performance of the proposed fusion approach}
\label{appendix:1_time_performance}

\begin{table*}[!ht]
\centering
\resizebox{\textwidth}{!}{%
\begin{tabular}{@{}ccc|c|cccccc|ccccccc@{}}
\toprule
\multirow{2}{*}{\textbf{Dataset}} & \multirow{2}{*}{\textbf{CNN}} & \multirow{2}{*}{\textbf{\begin{tabular}[c]{@{}c@{}}Base CNN\\ training \\ time (m)\end{tabular}}} & \multirow{2}{*}{\textbf{\begin{tabular}[c]{@{}c@{}}Feature\\ extraction \\ time (m)\end{tabular}}} & \multicolumn{6}{c|}{\textbf{\begin{tabular}[c]{@{}c@{}}Statistical features based\\ classifier training time (m)\end{tabular}}} & \multicolumn{7}{c}{\textbf{Ensemble training time (m)}} \\ \cmidrule(l){5-17} 
 &  &  &  & \textbf{KNN} & \textbf{LDA} & \textbf{LR} & \textbf{RF} & \textbf{\begin{tabular}[c]{@{}c@{}}SVM\\ rbf\end{tabular}} & \textbf{\begin{tabular}[c]{@{}c@{}}SVM\\ sig\end{tabular}} & \textbf{Avg.} & \textbf{KNN} & \textbf{LDA} & \textbf{LR} & \textbf{RF} & \textbf{\begin{tabular}[c]{@{}c@{}}SVM\\ rbf\end{tabular}} & \textbf{\begin{tabular}[c]{@{}c@{}}SVM\\ sig\end{tabular}} \\ \midrule
\multirow{5}{*}{\textbf{\begin{tabular}[c]{@{}c@{}}Caltech \\ birds\\ 2011\end{tabular}}} & \textbf{DenseNet201} & 66,00 & \multirow{5}{*}{0,77} & \multirow{5}{*}{2,45} & \multirow{5}{*}{2,51} & \multirow{5}{*}{2,49} & \multirow{5}{*}{2,47} & \multirow{5}{*}{2,48} & \multirow{5}{*}{2,46} & \multirow{5}{*}{2,02E-04} & \multirow{5}{*}{5,40} & \multirow{5}{*}{0,96} & \multirow{5}{*}{3,25} & \multirow{5}{*}{18,00} & \multirow{5}{*}{106,68} & \multirow{5}{*}{148,98} \\
 & \textbf{Inc. ResNet V2} & 52,60 &  &  &  &  &  &  &  &  &  &  &  &  &  &  \\
 & \textbf{Inc. V3} & 43,20 &  &  &  &  &  &  &  &  &  &  &  &  &  &  \\
 & \textbf{VGG19} & 20,25 &  &  &  &  &  &  &  &  &  &  &  &  &  &  \\
 & \textbf{Xception V1} & 35,40 &  &  &  &  &  &  &  &  &  &  &  &  &  &  \\ \midrule
\multirow{5}{*}{\textbf{\begin{tabular}[c]{@{}c@{}}Cars\\ 196\end{tabular}}} & \textbf{DenseNet201} & 116,40 & \multirow{5}{*}{1,73} & \multirow{5}{*}{4,65} & \multirow{5}{*}{4,60} & \multirow{5}{*}{4,62} & \multirow{5}{*}{4,60} & \multirow{5}{*}{4,62} & \multirow{5}{*}{4,67} & \multirow{5}{*}{3,66E-04} & \multirow{5}{*}{9,67} & \multirow{5}{*}{1,09} & \multirow{5}{*}{4,36} & \multirow{5}{*}{6,56} & \multirow{5}{*}{151,83} & \multirow{5}{*}{114,61} \\
 & \textbf{Inc. ResNet V2} & 93,00 &  &  &  &  &  &  &  &  &  &  &  &  &  &  \\
 & \textbf{Inc. V3} & 68,60 &  &  &  &  &  &  &  &  &  &  &  &  &  &  \\
 & \textbf{VGG19} & 21,25 &  &  &  &  &  &  &  &  &  &  &  &  &  &  \\
 & \textbf{Xception V1} & 77,40 &  &  &  &  &  &  &  &  &  &  &  &  &  &  \\ \midrule
\multirow{5}{*}{\textbf{Cassava}} & \textbf{DenseNet201} & 81,20 & \multirow{5}{*}{0,81} & \multirow{5}{*}{0,86} & \multirow{5}{*}{0,86} & \multirow{5}{*}{0,87} & \multirow{5}{*}{0,87} & \multirow{5}{*}{0,86} & \multirow{5}{*}{0,88} & \multirow{5}{*}{8,02E-06} & \multirow{5}{*}{1,89} & \multirow{5}{*}{0,09} & \multirow{5}{*}{0,14} & \multirow{5}{*}{0,60} & \multirow{5}{*}{0,40} & \multirow{5}{*}{0,54} \\
 & \textbf{Inc. ResNet V2} & 55,00 &  &  &  &  &  &  &  &  &  &  &  &  &  &  \\
 & \textbf{Inc. V3} & 45,00 &  &  &  &  &  &  &  &  &  &  &  &  &  &  \\
 & \textbf{VGG19} & 45,00 &  &  &  &  &  &  &  &  &  &  &  &  &  &  \\
 & \textbf{Xception V1} & 35,80 &  &  &  &  &  &  &  &  &  &  &  &  &  &  \\ \midrule
\multirow{5}{*}{\textbf{\begin{tabular}[c]{@{}c@{}}Cats vs\\ dogs\end{tabular}}} & \textbf{DenseNet201} & 72,40 & \multirow{5}{*}{1,43} & \multirow{5}{*}{5,67} & \multirow{5}{*}{5,66} & \multirow{5}{*}{5,64} & \multirow{5}{*}{5,57} & \multirow{5}{*}{5,66} & \multirow{5}{*}{5,63} & \multirow{5}{*}{1,89E-05} & \multirow{5}{*}{1,51} & \multirow{5}{*}{0,06} & \multirow{5}{*}{0,07} & \multirow{5}{*}{0,66} & \multirow{5}{*}{0,26} & \multirow{5}{*}{0,49} \\
 & \textbf{Inc. ResNet V2} & 68,80 &  &  &  &  &  &  &  &  &  &  &  &  &  &  \\
 & \textbf{Inc. V3} & 71,40 &  &  &  &  &  &  &  &  &  &  &  &  &  &  \\
 & \textbf{VGG19} & 92,00 &  &  &  &  &  &  &  &  &  &  &  &  &  &  \\
 & \textbf{Xception V1} & 67,40 &  &  &  &  &  &  &  &  &  &  &  &  &  &  \\ \midrule
\multirow{5}{*}{\textbf{\begin{tabular}[c]{@{}c@{}}Citrus\\ leaves\end{tabular}}} & \textbf{DenseNet201} & 40,20 & \multirow{5}{*}{0,06} & \multirow{5}{*}{0,02} & \multirow{5}{*}{0,02} & \multirow{5}{*}{0,02} & \multirow{5}{*}{0,02} & \multirow{5}{*}{0,02} & \multirow{5}{*}{0,02} & \multirow{5}{*}{1,47E-06} & \multirow{5}{*}{0,11} & \multirow{5}{*}{0,02} & \multirow{5}{*}{0,03} & \multirow{5}{*}{0,59} & \multirow{5}{*}{0,02} & \multirow{5}{*}{0,03} \\
 & \textbf{Inc. ResNet V2} & 17,00 &  &  &  &  &  &  &  &  &  &  &  &  &  &  \\
 & \textbf{Inc. V3} & 19,60 &  &  &  &  &  &  &  &  &  &  &  &  &  &  \\
 & \textbf{VGG19} & 7,50 &  &  &  &  &  &  &  &  &  &  &  &  &  &  \\
 & \textbf{Xception V1} & 15,20 &  &  &  &  &  &  &  &  &  &  &  &  &  &  \\ \midrule
\multirow{5}{*}{\textbf{\begin{tabular}[c]{@{}c@{}}Deep\\ weeds\end{tabular}}} & \textbf{DenseNet201} & 118,80 & \multirow{5}{*}{0,97} & \multirow{5}{*}{3,26} & \multirow{5}{*}{3,27} & \multirow{5}{*}{3,26} & \multirow{5}{*}{3,27} & \multirow{5}{*}{3,27} & \multirow{5}{*}{3,24} & \multirow{5}{*}{2,54E-05} & \multirow{5}{*}{12,49} & \multirow{5}{*}{0,17} & \multirow{5}{*}{0,31} & \multirow{5}{*}{0,63} & \multirow{5}{*}{1,04} & \multirow{5}{*}{1,17} \\
 & \textbf{Inc. ResNet V2} & 127,40 &  &  &  &  &  &  &  &  &  &  &  &  &  &  \\
 & \textbf{Inc. V3} & 83,60 &  &  &  &  &  &  &  &  &  &  &  &  &  &  \\
 & \textbf{VGG19} & 93,75 &  &  &  &  &  &  &  &  &  &  &  &  &  &  \\
 & \textbf{Xception V1} & 54,40 &  &  &  &  &  &  &  &  &  &  &  &  &  &  \\ \midrule
\multirow{5}{*}{\textbf{Malaria}} & \textbf{DenseNet201} & 95,20 & \multirow{5}{*}{1,27} & \multirow{5}{*}{6,22} & \multirow{5}{*}{6,11} & \multirow{5}{*}{6,25} & \multirow{5}{*}{6,26} & \multirow{5}{*}{6,20} & \multirow{5}{*}{6,11} & \multirow{5}{*}{1,93E-05} & \multirow{5}{*}{1,80} & \multirow{5}{*}{0,10} & \multirow{5}{*}{0,07} & \multirow{5}{*}{0,70} & \multirow{5}{*}{0,28} & \multirow{5}{*}{0,65} \\
 & \textbf{Inc. ResNet V2} & 104,80 &  &  &  &  &  &  &  &  &  &  &  &  &  &  \\
 & \textbf{Inc. V3} & 89,40 &  &  &  &  &  &  &  &  &  &  &  &  &  &  \\
 & \textbf{VGG19} & 68,50 &  &  &  &  &  &  &  &  &  &  &  &  &  &  \\
 & \textbf{Xception V1} & 74,60 &  &  &  &  &  &  &  &  &  &  &  &  &  &  \\ \midrule
\multirow{5}{*}{\textbf{\begin{tabular}[c]{@{}c@{}}Oxford\\ flowers\\ 102\end{tabular}}} & \textbf{DenseNet201} & 45,40 & \multirow{5}{*}{0,72} & \multirow{5}{*}{0,17} & \multirow{5}{*}{0,17} & \multirow{5}{*}{0,17} & \multirow{5}{*}{0,17} & \multirow{5}{*}{0,17} & \multirow{5}{*}{0,17} & \multirow{5}{*}{6,54E-05} & \multirow{5}{*}{1,21} & \multirow{5}{*}{0,29} & \multirow{5}{*}{0,53} & \multirow{5}{*}{1,20} & \multirow{5}{*}{9,69} & \multirow{5}{*}{10,08} \\
 & \textbf{Inc. ResNet V2} & 24,40 &  &  &  &  &  &  &  &  &  &  &  &  &  &  \\
 & \textbf{Inc. V3} & 25,00 &  &  &  &  &  &  &  &  &  &  &  &  &  &  \\
 & \textbf{VGG19} & 18,50 &  &  &  &  &  &  &  &  &  &  &  &  &  &  \\
 & \textbf{Xception V1} & 23,60 &  &  &  &  &  &  &  &  &  &  &  &  &  &  \\ \midrule
\multirow{5}{*}{\textbf{\begin{tabular}[c]{@{}c@{}}Plant\\ leaves\end{tabular}}} & \textbf{DenseNet201} & 133,20 & \multirow{5}{*}{15,09} & \multirow{5}{*}{0,32} & \multirow{5}{*}{0,32} & \multirow{5}{*}{0,32} & \multirow{5}{*}{0,32} & \multirow{5}{*}{0,32} & \multirow{5}{*}{0,32} & \multirow{5}{*}{1,63E-05} & \multirow{5}{*}{1,22} & \multirow{5}{*}{0,10} & \multirow{5}{*}{0,17} & \multirow{5}{*}{0,61} & \multirow{5}{*}{1,12} & \multirow{5}{*}{1,28} \\
 & \textbf{Inc. ResNet V2} & 68,20 &  &  &  &  &  &  &  &  &  &  &  &  &  &  \\
 & \textbf{Inc. V3} & 75,60 &  &  &  &  &  &  &  &  &  &  &  &  &  &  \\
 & \textbf{VGG19} & 65,00 &  &  &  &  &  &  &  &  &  &  &  &  &  &  \\
 & \textbf{Xception V1} & 58,60 &  &  &  &  &  &  &  &  &  &  &  &  &  &  \\ \midrule
\multirow{5}{*}{\textbf{\begin{tabular}[c]{@{}c@{}}Plant\\ village\end{tabular}}} & \textbf{DenseNet201} & 232,80 & \multirow{5}{*}{2,75} & \multirow{5}{*}{37,84} & \multirow{5}{*}{37,86} & \multirow{5}{*}{38,92} & \multirow{5}{*}{38,17} & \multirow{5}{*}{37,72} & \multirow{5}{*}{37,94} & \multirow{5}{*}{2,04E-04} & \multirow{5}{*}{130,35} & \multirow{5}{*}{0,77} & \multirow{5}{*}{3,12} & \multirow{5}{*}{3,02} & \multirow{5}{*}{36,56} & \multirow{5}{*}{44,47} \\
 & \textbf{Inc. ResNet V2} & 293,20 &  &  &  &  &  &  &  &  &  &  &  &  &  &  \\
 & \textbf{Inc. V3} & 212,80 &  &  &  &  &  &  &  &  &  &  &  &  &  &  \\
 & \textbf{VGG19} & 301,25 &  &  &  &  &  &  &  &  &  &  &  &  &  &  \\
 & \textbf{Xception V1} & 228,40 &  &  &  &  &  &  &  &  &  &  &  &  &  &  \\ \bottomrule
\end{tabular}%
}
\caption{Duration in minutes of the different parts of the approach proposed for the different datasets, CNN architectures, classifiers used for building the statistical features based classifier and algorithms used for the final ensemble.}
\label{tab:training_times}
\end{table*}

\end{document}